\documentclass[lettersize,journal]{IEEEtran}

\usepackage{hyperref}
\usepackage{graphicx} 
\usepackage{amsmath} 
\usepackage{amssymb}  
\usepackage{tikz}
\usetikzlibrary{positioning}
\usepackage{subcaption}
\usepackage{makecell}
\usepackage{float}
\usepackage{rotating}
\usepackage{cite}
\usepackage{caption}
\usepackage{pgfplots}
\usepackage{tabularx, booktabs}
\usepackage{textgreek}
\usepackage{tcolorbox}
\usepackage{subcaption}
\usepackage[linesnumbered,ruled,vlined,Algorithm ]{algorithm }
\usepackage{algpseudocode}
\usepackage{booktabs}
\usepackage{multirow}
\usepackage{siunitx}
\usepackage{hyperref}
\usepackage{tabularx}
\usepackage{adjustbox}
\usepackage{wrapfig}

\makeatletter
\let\NAT@parse\undefined
\makeatother
\DeclareMathSymbol{\shortminus}{\mathbin}{AMSa}{"39}\newcolumntype{C}{>{\centering\arraybackslash}X}

\title{An efficient Deep Spatio-Temporal Context Aware decision Network (DST-CAN) for Predictive Manoeuvre Planning on Highways}

\author{Jayabrata Chowdhury$^{1}$, Suresh Sundaram$^{2}$, Nishanth Rao$^{3}$, Narasimhan Sundararajan$^{4}$ 
\thanks{$^{1}$Jayabrata Chowdhury is with Robert Bosch Centre for Cyber-Physical Systems \& Artificial Intelligence and Robotics Lab, Dept. of Aerospace Engineering, Indian Institute of Science, Bangalore, India. $^{2}$Suresh Sundaram and $^{3}$Nishanth Rao are with the Artificial Intelligence and Robotics Lab, Dept. of Aerospace Engineering, Indian Institute of Science, Bangalore, India. $^{4}$Narasimhan Sundararajan is a Technical Consultant, WIRIN project, Dept. of Aerospace Engineering, Indian Institute of Science, Bangalore, India. {\tt\small Email: (jayabratac,vssuresh,nishanthrao)@iisc.ac.in and ensundara@gmail.com}}
\thanks{Manuscript received April 5, 2022}
}

\begin{document}
\tikzset{%
	every neuron/.style={
		circle, 
		draw,
		minimum size = 0.7cm
	},
	neuron memory/.style={
		circle,
		draw,
		minimum size = 0.5cm, 
		fill=black
	},
	neuron missing/.style={
		draw=none,
		scale=2.5,
		text height=0.133cm,
		execute at begin node=\color{black}$\vdots$
	},
}
\markboth{Journal of \LaTeX\ Class Files,~Vol.~14, No.~8, August~2021}%
{Shell \MakeLowercase{\textit{et al.}}: A Sample Article Using IEEEtran.cls for IEEE Journals}


\maketitle

\begin{abstract}
The safety and efficiency of an Autonomous Vehicle (AV) manoeuvre planning heavily depend on the future trajectories of surrounding vehicles. If an AV can predict its surrounding vehicles' future trajectories, it can make safe and efficient manoeuvre decisions. In this paper, we present a Deep Spatio-Temporal Context-Aware decision Network (DST-CAN) for predictive manoeuvre decisions for AVs on highways. DST-CAN has two main components, namely spatio-temporal context-aware map generator and predictive manoeuvre decisions engine. DST-CAN employ a memory neuron network to predict the future trajectories of its surrounding vehicles. Using look-ahead prediction and past actual trajectories, a spatio-temporal context-aware probability occupancy map is generated. These context-aware maps as input to a decision engine generate a safe and efficient manoeuvre decision. Here, CNN helps extract feature space, and two fully connected network generates longitudinal and lateral manoeuvre decisions. Performance evaluation of DST-CAN has been carried out using two publicly available NGSIM US-101 and I-80 highway datasets. A traffic rule is defined to generate ground truths for these datasets in addition to human decisions. Two DST-CAN models are trained using imitation learning with human driving decisions from actual traffic data and rule-based ground truth decisions. The performances of the DST-CAN models are compared with the state-of-the-art Convolutional Social-LSTM (CS-LSTM) models for manoeuvre prediction. The results clearly indicate that the context-aware maps help DST-CAN to predict the decision accurately over CS-LSTM. Further, an ablation study has been carried out to understand the effect of prediction horizons of performance and a robustness study to understand the near collision scenarios over actual traffic observations. The context-aware map with a $3$ second prediction horizon is most suitable and robust against near collision. 
\end{abstract}


\section{Introduction}
\IEEEPARstart{I}{}t is very crucial for an Autonomous Vehicle (AV) to understand the intentions of its surrounding vehicles based on their past and present contextual information for safe and efficient decision-making. These intentions can be inferred by predicting the future trajectories of surrounding vehicles. However, the dynamic nature of the context of the AV and the uncertainties in the prediction \cite{katariya2022Deep_track} of surrounding vehicles' trajectories pose a significant challenge in manoeuvre planning that may lead to near-collision scenarios. A Predictive Manoeuvre Planning (PMP) approach that can incorporate the contextual information with uncertainties in predicting neighbouring vehicles' trajectories is crucial for safe and efficient manoeuvre planning. 

In PMP, it is assumed that the perception module provides the relative positions of surrounding vehicles for input to the trajectory prediction module. Recurrent Neural Networks (RNNs) have been a popular choice for trajectory prediction. In particular, Long Short-Term Memory (LSTM) networks have been used in previous works such as \cite{Park2018MapPrediction}, \cite{Hou2020LSTMprediction}, while Bidirectional Gated Recurrent Unit (Bi-GRU) networks have been used in \cite{shu2022short_term_traffic}. However, these works only model the temporal relationships between vehicles and do not account for the spatial relationships among them. More recently, a method has been proposed to combine both the spatial and temporal relationships for trajectory prediction in \cite{katariya2022Deep_track}. However, a more accurate and efficient trajectory prediction module has been developed in \cite{rao2021spatio} by formulating the problem as a sequence generation problem and using a recurrent Memory Neuron Network (MNN) \cite{sastry1994memory} to learn the functional relationship between the past and future trajectory samples. Despite these advances, all of these works still need to address the issue of decision-making using predicted trajectories for safe manoeuvre planning in the presence of prediction uncertainties.

Another important problem in PMP is to generate safe and efficient manoeuvre decisions from the predicted trajectories with uncertainties \cite{Chitta2023Transfuser, wu2022trajectory}. In particular, the lane change decision significantly contributes for the near-collision scenario in AV. In \cite{Lee2017LaneChangeIntention}, a Convolutional Neural Network (CNN) was used to identify the lane change decision of surrounding vehicles and model predictive controller generate the trajectory for safe and comfortable rides for passengers in AV. The past and present position alone is not sufficient to accurately determine the lateral decision using CNN. Recently, in \cite{deo2018convolutional}, the spatial and temporal relationships among the surrounding vehicles are combined for decision-making. This approach uses an LSTM network to build a social tensor encoding the past trajectory information of the ego and surrounding vehicles. Since this social tensor has spatio-temporal relationships, a CNN was used to extract the spatial interaction features of the vehicles to identify the lateral and longitudinal manoeuvre decisions. In \cite{Mersch2021spatio}, a CNN was used to find the correlations between surrounding vehicles' spatial and temporal information. However, all of these works only used past information for manoeuvre prediction and did not consider the future intentions of the neighbouring vehicles. In some recent autonomous driving frameworks (e.g., Apollo \cite{Fan2018Apollo}, Autoware \cite{Kato2018Autoware}), a prediction module has been incorporated for decision-making. However, these works use computationally intensive High-Definition (HD) maps for carrying out the predictions. Hence, one needs to predict the future trajectories of neighbouring vehicles from past observations and use predicted spatio-temporal observation to determine efficient and safe manoeuvers.

In this paper, a novel Deep Spatio-Temporal Context-Aware decision Network (called DST-CAN) is proposed for safe, efficient predictive manoeuvre planning. DST-CAN has two primary modules: a spatio-temporal context-aware map generator module, and a decision engine module. DST-CAN assume that a state-of-the-art perception module (like \cite{Voigtlaender_2019_MOTS, Baser_2019_FanTrack, Luo_2021_ICCV, Zhang_2019_sensor_fusion}) will provide the past trajectories of the surrounding vehicles using camera-based dynamic object tracking and segmentation. Next, the spatio-temporal context map generator module contains a pretrained MNN for predicting the future trajectories of surrounding vehicles for time horizon $T$ and a context-based Probabilistic Occupancy Map (POM) encoder for context-aware map generation. With an increase in time horizon, the prediction uncertainty also increases. Hence, a local POM is created from these predicted trajectories based on the error in prediction. The past occupancy grid maps and the predicted POMs are then combined to form spatio-temporal context-aware grid maps. Finally, the decision engine module consists of a CNN and two fully connected neural networks for lateral and longitudinal decisions. The CNN takes these maps as an input, generating an efficient feature space. The fully connected networks approximate the functional relationship between feature and decision space. The imitation learning approach with both human decisions and traffic rule-based ground truth decisions is used to train the DST-CAN. 

The publicly available NGSIM highway traffic dataset \cite{colyar2007us101} has been used for a performance evaluation of DST-CAN. For performance analysis, we have generated rule-based ground truth decisions based on the traffic rules (for safety and efficiency) and compared the results with ground truth decisions and human decisions from actual traffic data. Two models of the DST-CAN have been trained based on both human-driving decisions and rule-based decisions. From a study on different prediction horizons, it has been found that a $3$-second prediction horizon gives better results than a $1$ or $5$-second prediction horizon. Detailed simulation studies have been carried out for the performance evaluation of DST-CAN, and the results have been compared with the State-of-The-Art (SOTA) Convolutional Social Long-Short Term Memory (CS-LSTM) \cite{deo2018convolutional} model. Results with the conflict scenario show significant improvement in performance over CS-LSTM (improvement of $21.15 \%$ and $12.12 \%$ for lateral and longitudinal cases, respectively), whereas the non-conflict scenario DST-CAN is slightly better than CS-LSTM. The case study on near-collision scenarios shows that DST-CAN provides efficient manoeuvre decisions and, thereby, minimizes near-collision scenarios over traffic observation. Since trajectory prediction is a crucial component in the DST-CAN model, a detailed comparison has been made with recent unimodal and multimodal SOTA methods for trajectory prediction. Since DST-CAN is an unimodal method, results are compared with unimodal trajectory prediction only. A detailed analysis of manoeuvre prediction with the SOTA multimodal method can be found in the supplementary material.
\begin{figure}[h!]
\centering
\includegraphics[width=3.4in]{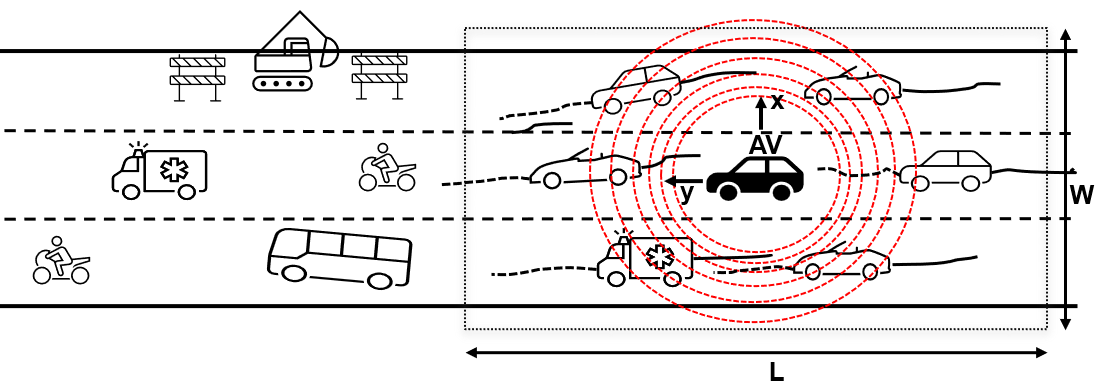}
\caption{A typical traffic situation with an AV in operation}
\label{figure1}
\end{figure}
The main contributions of the proposed DST-CAN are as follows:
\begin{itemize}
    \item DST-CAN models the evolving nature of a dynamic environment by learning the interplay between spatial features (like object locations) and temporal dependencies (how these features change over time). This allows the model to predict future states and react proactively and accurately to environmental changes
    \item Predictive uncertainty estimation is integrated into probabilistic occupancy map generation to guarantee safe navigation, enabling agents to prioritize manoeuvre and avoid areas with high predicted future occupancy.
    \item We introduce a novel data-pruning approach tailored explicitly for long-tail datasets to address the challenges of learning from imbalanced data in imitation tasks.
\end{itemize}

The paper is organized as follows. Section \ref{PMP Problem Description} gives a broad overview of the PMP problem. Section \ref{Deep Spatio-Temporal Context-Aware decision Network (DST-CAN) for Predictive Manoeuvre Planning} describes the proposed DST-CAN approach, including the procedure for creating a spatio-temporal context-aware grid map and  CNN-based decision engine for safe and efficient manoeuvres. Section \ref{Performance Evaluation of DST-CAN} details public traffic datasets used in this study, the simulation setup and finally, detailed performance comparison results. Section \ref{CONCLUSIONS} summarizes the study's main conclusions.

\section{PMP Problem Formulation} \label{PMP Problem Description}
\begin{figure*}[h!]
    \begin{tikzpicture}
        \draw (0,0) rectangle (1.8,1.8);
        \node at (0.9, 1.3) {AV};
        \node at (0.9, 0.9) {Perception};
        \node at (0.9, 0.5) {Engine};
        
        \draw[-latex, ultra thick] (1.9, 0.9) -- (3.45, 0.9);
        \node at (2.4, 1.4) {$\begin{pmatrix} \pmb{P_a} \\ \pmb{P_o} \end{pmatrix}$};
        
        \draw (3.5, 0) rectangle (5.5, 1.8);
        \node at (4.5, 1.3) {MNN};
        \node at (4.5, 0.9) {Look-ahead};
        \node at (4.5, 0.5) {Predictor};
        
        \draw[-latex, ultra thick] (5.6, 0.9) -- (7.25, 0.9); 
        \node at (6.3, 1.9) {$\begin{pmatrix} \pmb{P_a} \\ \pmb{P_o} \end{pmatrix}$};
        \node at (6.3, 1.25) {past};
        
        \draw[-latex, thick] (2.4, 1.9) -- (2.4, 2.8) -- (6.3, 2.8) -- (6.3, 2.32);
        
        \node at (6.3, 0.45) {$\left( \pmb{\hat{P_o}} \right)$};
        \node at (6.3, -0.1) {predicted};
        \node at (6.3, -0.35) {future};
        
        \draw (7.3, 0) rectangle (9.3, 1.8);
        \node at (8.3, 1.3) {Context};
        \node at (8.3, 0.9) {based POM};
        \node at (8.3, 0.5) {Encoder};
        
        \draw[dashed] (3.0, -0.65) rectangle (9.8, 2.55);
        
        \node at (6.3, -1.0) {\textbf{Spatio-temporal context-aware map generator}};
        
        \draw[-latex, ultra thick] (9.4, 0.9) -- (10.65, 0.9);
        
        \draw (11, 0) rectangle (12, 1.8);
        \draw (10.7, 0.2) rectangle (11.7, 2.0);
        \draw (11.3, -0.2) rectangle (12.3, 1.6);
        
        \node at (11.3, 2.8) {\small Context \& POM};
        \node at (11.3, 2.45) {\small grid maps};
        
        \draw[-latex, ultra thick] (12.4, 0.9) -- (13.45, 0.9);
        
        \draw (13.5, -0.2) rectangle (15.9, 2.0);
        \node at (14.7, 1.8) {Convolutional};
        \node at (14.7, 1.4) {Neural Net};
        \node at (14.7, 1.1) {+};
        \node at (14.7, 0.8) {2 Fully};
        \node at (14.7, 0.5) {Connected};
        \node at (14.7, 0.1) {Network};
        
        \draw[dashed] (10.3, -0.5) rectangle (16.2, 2.2);
        
        \node at (13.5, -1.0) {\textbf{Predictive manoeuvre decisions engine}};
        
        \draw[-latex, ultra thick] (16, 1.6) -- (16.7, 1.6);
        \draw[-latex, ultra thick] (16, 0.2) -- (16.7, 0.2);
        
        \node at (17.2, 1.6) {$\begin{Bmatrix} S \\ L \\ R \end{Bmatrix}$};
        \node at (17.2, 0.2) {$\begin{Bmatrix} C \\ B \end{Bmatrix}$};
        
    \end{tikzpicture}
\caption{A schematic diagram for PMP using  \textbf{D}eep \textbf{S}patio \textbf{T}emporal \textbf{C}ontext-\textbf{A}ware decision \textbf{N}etwork (\textbf{DST-CAN})}
\label{figure3}
\end{figure*}
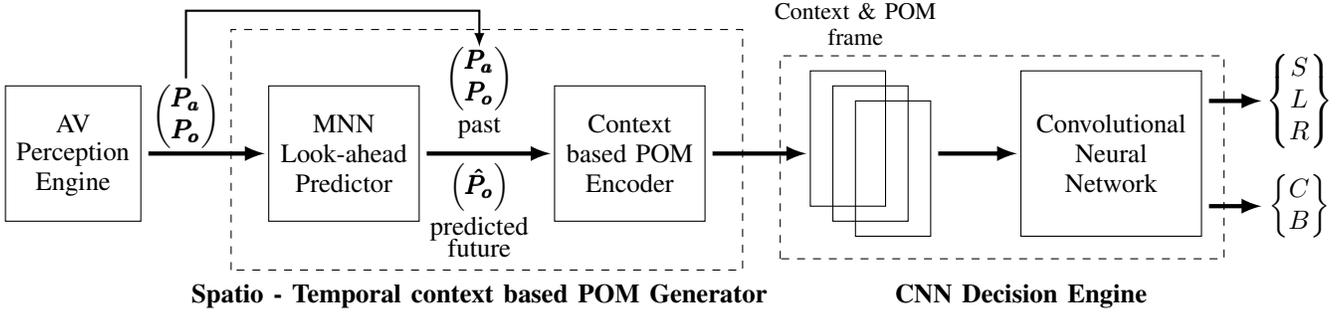
One of the crucial challenges for operating a vehicle in a natural environment is to develop its capability to understand the intentions of its surrounding vehicles and then plan its safe and efficient manoeuvres. To understand the importance of the prediction of the neighbouring vehicle's intentions, a typical traffic scenario with an AV in operation (at any given time instant (t) is shown in Fig. \ref{figure1} for illustration. The red dotted circular lines indicate the AV's sensing region. AV cannot sense the static obstacles present in the top row, which influences the traffic on the left lane. Based on the current and past positions of static/dynamic objects, the AV needs to anticipate the neighbours' intentions with predicted trajectories (as contextual information) and plan its manoeuvre accordingly. The manoeuvre decision should be safe (without colliding with other vehicles), follow the standard traffic rules, and also be efficient. The problem of manoeuvre planning (i.e., action taken by the AV without colliding with other objects) by understanding the intentions of its neighbours is referred to as the Predictive Manoeuvre Planning (PMP) problem. It is assumed that the AV can detect the relative positions of both static and dynamic objects. Thus, the AV has a complete history of paths traversed by the dynamic objects present at any given time and is also aware of the local map. The PMP problem is then formulated based on the histories of the paths traversed by the dynamic objects present in the scene and also the local map.

With the above aim, a rectangular window of length $L$ and width $W$ with the AV at the centre is considered, as shown in Fig.\ref{figure1}. The past trajectories of the vehicles are represented with dark lines, and the predicted look-ahead trajectories of the dynamic objects (non-ego motion vehicles) are represented using dashed lines. The figure shows the path traversed by AV for the past $n$ steps (${\textbf{P}_a}$) and the paths traversed by the other dynamic objects for the past $n$ steps seen at time $t$ (${\textbf{P}_o}$). In PMP, it is essential to predict the future look-ahead trajectories of the other dynamic objects (${\hat{\textbf{P}}_o}$) to understand the occupancy map into which the AV will move in the next $m$ steps in the near future. The ego vehicle's decision space ($A$) is divided into both longitudinal and lateral manoeuvre decisions. The two decisions for the longitudinal manoeuvre can be either 'cruise (C)' or 'brake (B).' The lateral manoeuvre action has three possibilities, and it can be either 'keep the same lane' (S) or 'change to the left lane' (L), or 'change to the right lane' (R). The actions should be safe, efficient, and follow traffic rules. The proposed DST-CAN method, currently designed for autonomous vehicles on highways, can be adapted for urban scenarios with suitable modifications to incorporate traffic rules and the driving map structure. Changes that are needed may include adjusting for different lane numbers and altering the context window length based on average vehicle speed and the road speed limits. The main objective of DST-CAN is to find the relationship between spatio-temporal context-aware maps and decision space. One can implement the decision sequence for actual trajectory control. For trajectory control for longitudinal and lateral manoeuvres, one can use the approach given in \cite{debarshi2022robust} such that the overall system is stable. It is important to perform a stability analysis before deploying the application.

\section{DST-CAN for Predictive Manoeuvre Planning}
\label{Deep Spatio-Temporal Context-Aware decision Network (DST-CAN) for Predictive Manoeuvre Planning}
The PMP problem considered in this paper consists of a trajectory prediction problem followed by a manoeuvre decision problem. For carrying out this, as a first step, one needs to encode the context information up to the current instant for the AV and then develop a model to predict the look-ahead occupancy map of the neighbouring dynamic obstacles. Next, one needs to approximate the functional relationships between the context information of the surrounding vehicles and the possible manoeuvre decision class for the AV to meet the desired objectives. 

\subsection{Deep Spatio-Temporal Context-Aware decision Network} 
The schematic diagram of the proposed DST-CAN using context-aware decision-making using the PMP model is given in Fig.\ref{figure3}. This figure shows that the overall scheme has three major modules: viz., a perception engine, a spatio-temporal context-aware map generator, and a predictive manoeuvre decisions engine block. The perception engine block of AV detects and tracks other objects. The context-aware map generator block generates the neighbouring vehicles' current context and future occupancy maps. As the final step, based on the above information, the decision engine predicts the manoeuvre decisions that follow the traffic rule for safe and efficient manoeuvres.
\begin{equation}
\begin{aligned}
    \{\textbf{P}_a\}=\{(x^{-n}_a,y^{-n}_a),(x^{-n+1}_a,y^{-n+1}_a),...,(x^0_a,y^0_a)\} \\
    \{\textbf{P}_o\}=\{(x^{-n}_{o_{i}},y^{-n}_{o_{i}}),(x^{-n+1}_{o_{i}},y^{-n+1}_{o_{i}}),...,(x^0_{o_{i}},y^0_{o_{i}})\} \\
    \{\hat{\textbf{P}}_o\}=\{(\hat{x}^{1}_{o_{i}},\hat{y}^{1}_{o_{i}}),(\hat{x}^{2}_{o_{i}},\hat{y}^{2}_{o_{i}}),...,(\hat{x}^m_{o_{i}},\hat{y}^m_{o_{i}})\} \\
    \text{where } i=1,2,...,D; \text{$D$ is number of dynamic objects}
    \label{eq:history}
\end{aligned}
\end{equation}
The perception engine in AV identifies the dynamic objects at the current instant and tracks the paths traversed by these objects, as shown in the first module of Fig.\ref{figure3}. However, in this work, real-world traffic data has been used instead of a real-time camera feed. This dataset has tracked vehicles' trajectories using object-tracking techniques. DST-CAN consists of a context-aware map generator using a look-ahead predictor and a CNN-based decision engine to relate the context-aware maps and manoeuvre decisions. The context-aware POM generator uses the paths traversed by the AV (${\textbf{P}_a}$) and dynamic objects (${\textbf{P}_o}$) up to time $t$ as given in Equ. \ref{eq:history} and generates their look-ahead trajectories (predictions) (${\hat{\textbf{P}}_o}$) using a MNN \cite{rao2021spatio, sastry1994memory}. A context encoder then encodes these predicted trajectories into a probability occupancy density map. The resultant context and occupancy maps are then used as inputs to the predictive manoeuvre decision engine to obtain the final manoeuvre planning decisions.

\subsubsection{Spatio-temporal Context-aware Map Generator}
The objective of the context and POM encoder is to generate a context and occupancy grid map in which the AV was present in the past. It also predicts the potential look-ahead trajectories of the other dynamic objects in the scene to understand the future intentions of these objects. The AV perception engine provides the path traversed by the AV (${\textbf{P}_a}$) for $n$-steps and the paths traversed by the dynamic objects up to the current time step (${\textbf{P}_o}$). Data-driven look-ahead prediction approaches have been widely used for prediction, such as recurrent neural networks \cite{rao2021spatio}, long-short term memory networks \cite{zyner2017long}, \cite{altche2017lstm}, and fuzzy neural networks \cite{subhrajit2020bayesian}. Recently, it was shown that the MNN-based approach \cite{rao2021spatio} is more effective for look-ahead trajectory prediction than other state-of-the-art approaches.

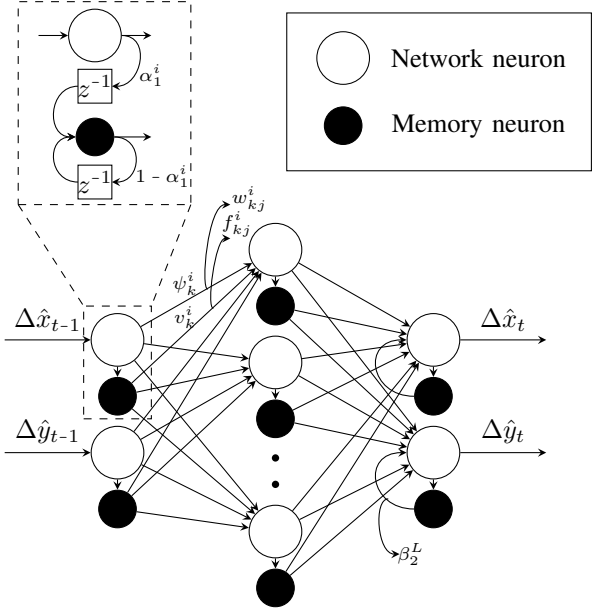
\begin{figure}
\begin{adjustbox}{max size={.95\textwidth}{.3\textheight}}
\begin{tikzpicture}[x=1.5cm, y=1.5cm, >=stealth]

\node [every neuron](nn-demo) at (0.8, 4.2) {};
\draw [<-] (nn-demo) -- ++(-0.5, 0);
\draw [->] (nn-demo) -- ++(0.5, 0);
\draw (0.65, 3.6) rectangle (0.95, 3.9) node[pos=.5]{$z^{\shortminus 1}$};
\node [neuron memory](mnn-demo) at (0.8, 3.3) {};
\draw [->] (mnn-demo) -- ++(0.5, 0);
\draw (0.65, 2.75) rectangle (0.95, 3.05) node[pos=.5]{$z^{\shortminus 1}$};
\draw [->] (nn-demo.east) .. controls +(left:-4mm) and +(right:4mm) .. (0.95, 3.75) node at (1.3, 3.85) {\scriptsize $\alpha_1^i$};
\draw [->] (0.65, 3.75) .. controls +(left:4mm) and +(right:-4mm) .. (mnn-demo.west);
\draw [->] (mnn-demo.east) .. controls +(left:-4mm) and +(right:4mm) .. (0.95, 2.9) node at (1.4, 2.95) {\scriptsize $1\shortminus\alpha_1^i$};
\draw [->] (0.65, 2.9) .. controls +(left:4mm) and +(right:-4mm) .. (mnn-demo.west);

\draw [dashed] (0.12, 2.7) -- (1.65, 2.7) -- (1.65, 4.5) -- (0.12, 4.5) -- cycle;
\draw [dashed] (0.12, 2.7) -- (0.7, 1.8);
\draw [dashed] (1.65, 2.7) -- (1.3,  1.8);

\node [every neuron](nn-legend) at (3.0, 4.0) {};
\node [neuron memory](mnn-legend) at (3.0, 3.4) {};
\node at (4.2, 4.0) {Network neuron}; 
\node at (4.2, 3.4) {Memory neuron};

\draw (2.5, 4.4) -- (5.2, 4.4) -- (5.2, 3.0) -- (2.5, 3.0) -- cycle;
\foreach \i in {1, 2}
	\node [every neuron/.try](input-\i) at (1.0,2.5-\i) {};

\foreach \i in {1, 2}
	\node [neuron memory/.try](input-memory-\i) at (1.0, 2.0-\i) {};

\foreach \l [count=\i] in {x, y}
	\draw [<-] (input-\i) -- ++(-1, 0)
		node [above, midway] {$\Delta\hat{\l}_{t\shortminus 1}$};
		
\foreach \i in {1, 2}
	\draw[->] (input-\i) -- (input-memory-\i);

\foreach \m [count=\i] in {1, memory, 3, memory, missing, 6, memory}
	\node [every neuron/.try, neuron \m/.try](hidden-\m-\i) at (2.4, 2.8-\i/2) {};

\foreach \i\j in {1/2, 3/4, 6/7}
	\draw[->] (hidden-\i-\i) -- (hidden-memory-\j);


\foreach \i in {1, 2}
	\node [every neuron/.try](output-\i) at (3.8,2.5-\i) {};

\foreach \i in {1, 2}
	\node [neuron memory/.try](output-memory-\i) at (3.8, 2.0-\i) {};

\foreach \i in {1, 2}
	\draw[->] (output-\i) -- (output-memory-\i);
	
\foreach \i in {1, 2}
	\foreach \j in {1, 3, 6}
		\draw [->] (input-\i) -- (hidden-\j-\j);

\foreach \i in {1, 2}
	\foreach \j in {1, 3, 6}
		\draw [->] (input-memory-\i) -- (hidden-\j-\j);

\foreach \i in {1, 3, 6}
	\foreach \j in {1, 2}
		\draw [->] (hidden-\i-\i) -- (output-\j);

\foreach \i in {2, 4, 7}
	\foreach \j in {1, 2}
		\draw [->] (hidden-memory-\i) -- (output-\j);

\foreach \i in {1, 2}
	\draw [->] (output-memory-\i.west) .. controls +(left:6mm) and +(right:-6mm) .. (output-\i.west);
	
\foreach \l [count=\i] in {x, y}
	\draw [->] (output-\i) -- ++(1, 0)
		node [above, midway] {$\Delta\hat{\l}_{t}$};

\node at (1.6, 2.0) {\scriptsize $\psi_k^i$};
\node at (1.6, 1.7) {\scriptsize $v_k^i$};
\draw [->] (1.8, 1.95) .. controls +(left:1mm) and +(right:-3mm) .. (2.0, 2.7);
\draw [->] (1.85, 1.78) .. controls +(left:1mm) and +(right:-2mm) .. (2.0, 2.4);

\node at (2.17, 2.75) {\scriptsize $w_{kj}^i$};
\node at (2.06, 2.53) {\scriptsize $f_{kj}^i$};

\draw [->] (3.4, 0.09) .. controls +(left:2mm) and +(right:-3mm) .. (3.5, -0.4);
\node at (3.6, -0.4) {\scriptsize $\beta_2^L$};

\draw [dashed] (0.7, 0.8) -- (1.3, 0.8) -- (1.3, 1.8) -- (0.7, 1.8) -- cycle;
\end{tikzpicture}
\end{adjustbox}
\centering
\caption{MNN architecture for look-ahead trajectory prediction} 
\label{MNN}
\end{figure}

Hence, an MNN has been used in this paper, and its training has been done with backpropagation through time similar to \cite{rao2021spatio} for look-ahead trajectory predictions. A series-parallel model is used for training, and a parallel-parallel model is used during testing to predict the look-ahead trajectories \cite{helicopter_dynamics}. From the paths traversed by the dynamic objects ($\{\bf{P}_o\}$), changes in displacements in both $x$ and $y$ directions are computed. The inputs to the MNN are $\Delta x(t-1)$ and $\Delta y(t-1)$, and the outputs of the network are $\Delta x(t)$ and $\Delta y(t)$. The MNN architecture used in the paper is shown in Fig.\ref{MNN}. Each layer has a network neuron (blank circle), and each network neuron has one memory neuron (dark circle). The net output $\phi_j^h(t)$ of the  $j^{th}$ network neuron in the hidden layer $h$ is:
\begin{equation}
    \phi^h_j(t) = g \left(\sum_{k=1}^2 w_{kj}^{i} \psi_k^i(t) + \sum_{k=1}^2 f_{kj}^i v_k^i(t) \right)
    \label{eq:mnn1}
\end{equation}
where, 
\begin{itemize}
    \item $\psi_k^i(t)$ is the output of the $k^{th}$ network neuron in the input layer $i$. In our case, $\psi_1^i(t) = \Delta{\hat{x}_{t-1}}$ and $\psi_2^i(t) = \Delta{\hat{y}_{t-1}}$. $(x,y)$ is position of a vehicle.
    \item $g(.) = tanh(.)$ is activation function of network neurons.
    \item $1 \leq h \leq H $. $H$ is number of network neurons in hidden layer. In this work $H=6$. Only one hidden layer has been used for this MNN.
\end{itemize}
The output of the memory neuron corresponding to the $j^{th}$ network neuron in the layer $l$ is given by:
\begin{equation}
    v_j^l(t) = \alpha_j^l \psi^l_j(t-1) + (1-\alpha^l_j)v^l_j(t-1)
\end{equation}
where the $\alpha$ is weight between each network neuron and memory neuron. The net output $n^L_j(t)$ of the $j^{th}$ network neuron in the last layer $L$ is calculated as:
\begin{equation}
    \phi^L_j(t) = g^L \left( \sum_{k=1}^H w^h_{kj} \psi^h_{kj}(t) + \sum_{k=1}^H f^h_{kj} v^h_k(t) + \beta ^L_j v^L_j(t) \right), 
    \label{eq:mnn3}
\end{equation}
where $1 \leq j \leq 2$. The inherent feedback connections in the memory neurons help in accurately capturing the dynamics. The network neurons in the hidden layer employ a bipolar sigmoidal activation function, and the output layer employs a linear activation function.
The Root Mean Square Error (RMSE) function for predicting future positions during training the MNN is:
\begin{equation}
    RMSE = \frac{\sum_{n=1}^{\mathcal{N}}\sqrt{\frac{\sum_{t=1}^{T_H}{\lVert \mathbf{x}_t^{(n)} - \hat{\mathbf{x}}_t^{(n)} \rVert}^2}{T_H}}}{\mathcal{N}} 
    \label{eq:MNN_LOSS}
\end{equation}
where $\mathbf{x}_t = \left[x(t), y(t)\right]$ and $\hat{\mathbf{x}}_t = \left[\hat{x}(t),\hat{y}(t)\right]$ are ground truth and predicted positions respectively. $\mathcal{N}$ is the number of vehicles at time \textit{t}. $T_H$ is the prediction time horizon. The network is trained using backpropagation through time algorithm as in \cite{sastry1994memory}. One can find more details on the parallel-parallel implementation used to predict the future $m$-time steps and a comparison with other state-of-the-art methods in \cite{rao2021spatio}.

Next, the workings of the spatio-temporal context and POM encoder are described in Algorithm \ref{DST-CAN code}. This encoder converts the trajectories of AV and the other dynamic objects in the scene into a context-aware grid. For this purpose, the encoder divides the road segment of length $L$ with AV at the centre into three tracks with equal widths. Further, it divides the road segment into $p$ equal parts. A spatial context map is developed for every past time step. Since the perception engine gives the distances to the other vehicles very accurately (using sensors like LiDAR), it can be assumed that distance estimation uncertainty is much smaller than each grid length. The grid value on the road segment is set to one if a vehicle occupies the grid. Otherwise, it is set to zero.
\begin{figure}[h]
\centering
\begin{subfigure}[b]{0.15\textwidth}
\centering
\includegraphics[scale=0.5]{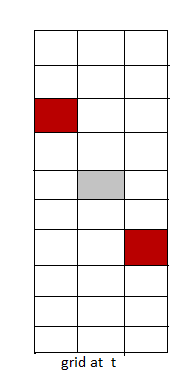}
\caption{context grid map}
\label{grid_t}
\end{subfigure}
\begin{subfigure}[b]{0.15\textwidth}
\centering
\includegraphics[scale=0.5]{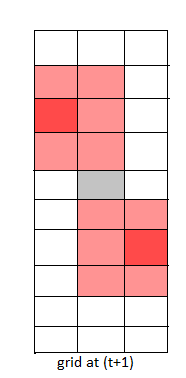}
\caption{POM}
\label{grid_t+1}
\end{subfigure}
\begin{subfigure}[b]{0.15\textwidth}
\centering
\includegraphics[scale=0.5]{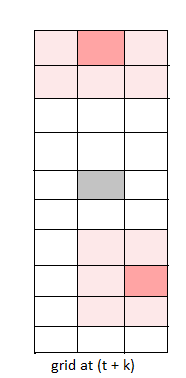}
\caption{POM}
\label{grid_t+k}
\end{subfigure}
\caption{Context and Probabilistic Occupancy Map (POM) embedding for different time steps}
\label{Context-POM-embedding}
\end{figure}
As a typical example, the spatial context map for the current instant is shown in Fig.\ref{grid_t}. The ego vehicle position is shown in grey colour. Dynamic objects are shown in red for occupancy; the darker the colour, the higher the occupancy probability. There is uncertainty in the predicted positions of surrounding vehicles. The RMSE for predictions is given in \cite{rao2021spatio}. RMSE for every look-ahead prediction step (seconds) has been normalized based on the max error observed to develop a function that can capture the uncertainty well. A decaying function of time has been fitted to relate the uncertainty of prediction to create the POM closely (uncertainty plot in supplementary material). This decaying function calculates how the certainty of prediction decreases with future time. This decaying function of time is calculated as given below:
\begin{equation}
    \textbf{P}(t) = 0.47 + \sqrt{0.236 - 0.004t}
    \label{eq:decay}
\end{equation}
where $0\le t \le 50$ is the time index.
\begin{figure*}
\begin{tikzpicture}
\draw (0+2,0+0.5) -- (0+2, 4.55+0.5) -- (1.5+2, 4.55+0.5) -- (1.5+2, 0+0.5) -- (0+2,0+0.5);
\foreach \i in {1, 2}
    \draw (0.5 * \i+2, 0+0.5) -- (0.5 * \i+2, 4.55+0.5);
\foreach \i in {1, 2, 3, 4, 5, 6, 7, 8, 9, 10, 11, 12}
    \draw (0+2, 0.35 * \i+0.5) -- (1.5+2, 0.35 * \i+0.5);
\fill [gray] (0.6+2, 2.1+0.5) rectangle (0.9+2, 2.45+0.5); 
\fill [red] (0.1+2, 3.15+0.5) rectangle (0.4+2, 3.5+0.5); 
\fill [red] (1.1+2, 1.05+0.5) rectangle (1.4+2, 1.4+0.5); 

\draw [thick] (0.75+2, 2.27+0.5) .. controls (0.8+2, 1.8+0.5) .. (0.75+2, 1.5+0.5); 
\draw [thick] (0.25+2, 3.32+0.5) .. controls (0.12+2, 2.7+0.5) .. (0.25+2, 2.55+0.5);
\draw [thick] (1.25+2, 1.22+0.5) .. controls (1.2+2, 0.8+0.5) .. (0.75+2, 0.52+0.5);


\draw [thick, dashed] (0.75+2, 2.27+0.5) .. controls (0.73 + 2, 2.6+0.5) .. (0.72+2, 3.04+0.5); 
\draw [thick, dashed] (0.25+2, 3.32+0.5) .. controls (0.38 + 2, 3.8+0.5) .. (0.75 + 2, 4.09+0.5);
\draw [thick, dashed] (1.25+2, 1.22+0.5) .. controls (1.3 + 2, 1.5+0.5) .. (1.25 + 2, 1.92+0.5);

\node at (2.5, 0.1) {occupancy map};
\node at (2.7, -0.2) {at $t=\tau$};

\draw (4.5, 0) -- (4.5, 4.55) -- (6.0, 5.416) -- (6.0, 0.866) -- cycle;
\foreach \i in {1, 2, 3, 4, 5, 6, 7, 8, 9, 10, 11, 12}
    \draw (4.5, 0.35 * \i) -- (6.0, 0.35 * \i + 0.866);
\foreach \i in {1, 2}
    \draw (4.5 + 0.5 * \i, 0.28867 * \i) -- (4.5 + 0.5 * \i, 4.55 + 0.28867 * \i);

\fill[cyan, opacity=0.15] (4.5,0) -- (4.5, 4.55) -- (10.5, 4.55) -- (10.5, 0) -- cycle;
\fill[cyan, opacity=0.15] (4.5,4.55) -- (6.0, 5.4160) -- (12.0, 5.4160) -- (10.5, 4.55) -- cycle;
\draw[cyan, opacity=0.8, dashed] (4.5, 4.55) -- (10.5, 4.55);
\draw[cyan, opacity=0.8, dashed] (4.5, 0) -- (10.5, 0);
\draw[cyan, opacity=0.8, dashed] (6.0, 5.4160) -- (12, 5.4160);

\fill [red] (5.0, 0.6386+0.7) -- (5.0, 0.9886+0.7) -- (5.5, 1.277 + 0.7) -- (5.5, 0.9272 + 0.7) -- cycle; 
\fill [red] (4.5, 2.1 + 0.35 + 0.7) -- (4.5, 2.45 + 0.35 + 0.7) -- (5.0, 2.7386 + 0.35 + 0.7) -- (5.0, 2.38867 + 0.35 + 0.7) -- cycle;

\draw (4.5 + 2, 0) -- (4.5 + 2, 4.55) -- (6.0 + 2, 5.416) -- (6.0 + 2, 0.866) -- cycle;
\foreach \i in {1, 2, 3, 4, 5, 6, 7, 8, 9, 10, 11, 12}
    \draw (4.5 + 2, 0.35 * \i) -- (6.0 + 2, 0.35 * \i + 0.866);
\foreach \i in {1, 2}
    \draw (4.5 + 2 + 0.5 * \i, 0.28867 * \i) -- (4.5 + 2 + 0.5 * \i, 4.55 + 0.28867 * \i);

\fill [red] (5.5 + 2, 1.6273) -- (5.5 + 2, 1.9773) -- (6 + 2, 2.266) -- (6 + 2, 1.916) -- cycle; 
\fill [red] (4.5 + 2, 2.1 + 0.35 + 0.7) -- (4.5 + 2, 2.45 + 0.35 + 0.7) -- (5.0 + 2, 2.7386 + 0.35 + 0.7) -- (5.0 + 2, 2.38867 + 0.35 + 0.7) -- cycle; 

\draw (4.5 + 4, 0) -- (4.5 + 4, 4.55) -- (6.0 + 4, 5.416) -- (6.0 + 4, 0.866) -- cycle;
\foreach \i in {1, 2, 3, 4, 5, 6, 7, 8, 9, 10, 11, 12}
    \draw (4.5 + 4, 0.35 * \i) -- (6.0 + 4, 0.35 * \i + 0.866);
\foreach \i in {1, 2}
    \draw (4.5 + 4 + 0.5 * \i, 0.28867 * \i) -- (4.5 + 4 + 0.5 * \i, 4.55 + 0.28867 * \i);

\fill [red, opacity=0.6] (5.5 + 4, 1.6273) -- (5.5 + 4, 1.9773) -- (6 + 4, 2.266) -- (6 + 4, 1.916) -- cycle; 
\fill [red, opacity=0.6] (4.5 + 4, 2.1 + 0.35 + 0.7) -- (4.5 + 4, 2.45 + 0.35 + 0.7) -- (5.0 + 4, 2.7386 + 0.35 + 0.7) -- (5.0 + 4, 2.38867 + 0.35 + 0.7) -- cycle; 
\fill [red, opacity=0.1] (5.5 + 4, 1.6273 - 0.35) -- (5.5 + 4, 1.9773-0.35) -- (6 + 4, 2.266-0.35) -- (6 + 4, 1.916-0.35) -- cycle;
\fill [red, opacity=0.1] (5.5 + 4, 1.6273 + 0.35) -- (5.5 + 4, 1.9773+0.35) -- (6 + 4, 2.266+0.35) -- (6 + 4, 1.916+0.35) -- cycle;
\fill [red, opacity=0.1] (9.0, 0.98867) -- (9.0, 1.33867) -- (9.5, 1.62734) -- (9.5, 1.27734) -- cycle;
\fill [red, opacity=0.1] (9.0, 0.98867 + 0.35) -- (9.0, 1.33867+ 0.35) -- (9.5, 1.62734+ 0.35) -- (9.5, 1.27734+ 0.35) -- cycle;
\fill [red, opacity=0.1] (9.0, 0.98867 + 0.7) -- (9.0, 1.33867 + 0.7) -- (9.5, 1.62734 + 0.7) -- (9.5, 1.27734 + 0.7) -- cycle;
\fill [red, opacity=0.1] (8.5, 3.15 + 0.35) -- (8.5, 3.5 + 0.35) -- (9.0, 3.78867 + 0.35) -- (9.0, 3.43867 + 0.35) -- cycle;
\fill [red, opacity=0.1] (8.5, 3.15 - 0.35) -- (8.5, 3.5 - 0.35) -- (9.0, 3.78867 - 0.35) -- (9.0, 3.43867 - 0.35) -- cycle;
\fill [red, opacity=0.1] (9.0, 3.43867) -- (9.0, 3.78867) -- (9.5, 4.07734) -- (9.5, 3.727) -- cycle;
\fill [red, opacity=0.1] (9.0, 3.43867 + 0.35) -- (9.0, 3.78867 + 0.35) -- (9.5, 4.07734 + 0.35) -- (9.5, 3.727 + 0.35) -- cycle;
\fill [red, opacity=0.1] (9.0, 3.43867 - 0.35) -- (9.0, 3.78867 - 0.35) -- (9.5, 4.07734 - 0.35) -- (9.5, 3.727 - 0.35) -- cycle;

\draw (4.5 + 6, 0) -- (4.5 + 6, 4.55) -- (6.0 + 6, 5.416) -- (6.0 + 6, 0.866) -- cycle;
\foreach \i in {1, 2, 3, 4, 5, 6, 7, 8, 9, 10, 11, 12}
    \draw (4.5 + 6, 0.35 * \i) -- (6.0 + 6, 0.35 * \i + 0.866);
\foreach \i in {1, 2}
    \draw (4.5 + 6 + 0.5 * \i, 0.28867 * \i) -- (4.5 + 6 + 0.5 * \i, 4.55 + 0.28867 * \i);

\fill [red, opacity=0.45] (5.5 + 6, 1.6273) -- (5.5 + 6, 1.9773) -- (6 + 6, 2.266) -- (6 + 6, 1.916) -- cycle; 
\fill [red, opacity=0.2] (5.5 + 6, 1.6273 - 0.35) -- (5.5 + 6, 1.9773-0.35) -- (6 + 6, 2.266-0.35) -- (6 + 6, 1.916-0.35) -- cycle;
\fill [red, opacity=0.2] (5.5 + 6, 1.6273 + 0.35) -- (5.5 + 6, 1.9773+0.35) -- (6 + 6, 2.266+0.35) -- (6 + 6, 1.916+0.35) -- cycle;
\fill [red, opacity=0.2] (9.0+2, 0.98867) -- (9.0+2, 1.33867) -- (9.5+2, 1.62734) -- (9.5+2, 1.27734) -- cycle;
\fill [red, opacity=0.2] (9.0+2, 0.98867 + 0.35) -- (9.0+2, 1.33867+ 0.35) -- (9.5+2, 1.62734+ 0.35) -- (9.5+2, 1.27734+ 0.35) -- cycle;
\fill [red, opacity=0.2] (9.0+2, 0.98867 + 0.7) -- (9.0+2, 1.33867 + 0.7) -- (9.5+2, 1.62734 + 0.7) -- (9.5+2, 1.27734 + 0.7) -- cycle;

\fill[red, opacity=0.45] (11, 3.78867) -- (11, 4.13867) -- (11.5, 4.42734) -- (11.5, 4.07734) -- cycle;
\fill[red, opacity=0.2] (11, 3.78867 + 0.35) -- (11, 4.13867 + 0.35) -- (11.5, 4.42734 + 0.35) -- (11.5, 4.07734 + 0.35) -- cycle;
\fill[red, opacity=0.2] (11, 3.78867 - 0.35) -- (11, 4.13867 - 0.35) -- (11.5, 4.42734 - 0.35) -- (11.5, 4.07734 - 0.35) -- cycle;

\fill[red, opacity=0.2] (11.5, 3.72734) -- (11.5, 4.07734) -- (12, 4.366) -- (12, 4.016) -- cycle;
\fill[red, opacity=0.2] (11.5, 3.72734 + 0.35) -- (11.5, 4.07734+ 0.35) -- (12, 4.366+ 0.35) -- (12, 4.016+ 0.35) -- cycle;
\fill[red, opacity=0.2] (11.5, 3.72734 + 0.7) -- (11.5, 4.07734 + 0.7) -- (12, 4.366 + 0.7) -- (12, 4.016 + 0.7) -- cycle;

\fill[red, opacity=0.2] (10.5, 3.5) -- (10.5, 3.85) -- (11, 4.13867) -- (11, 3.78867) -- cycle;
\fill[red, opacity=0.2] (10.5, 3.5+0.35) -- (10.5, 3.85+0.35) -- (11, 4.13867+0.35) -- (11, 3.78867+0.35) -- cycle;
\fill[red, opacity=0.2] (10.5, 3.5-0.35) -- (10.5, 3.85-0.35) -- (11, 4.13867-0.35) -- (11, 3.78867-0.35) -- cycle;

\fill[gray, opacity=0.3] (5.0, 2.38867) -- (5.0, 2.73867) -- (5.5, 3.0273) -- (5.5, 2.6773) -- cycle;
\fill[gray, opacity=0.3] (7.0, 2.38867) -- (7.0, 2.73867) -- (7.5, 3.0273) -- (7.5, 2.6773) -- cycle;
\fill[gray, opacity=0.3] (9.0, 2.38867) -- (9.0, 2.73867) -- (9.5, 3.0273) -- (9.5, 2.6773) -- cycle;
\fill[gray, opacity=0.3] (11.0, 2.38867) -- (11.0, 2.73867) -- (11.5, 3.0273) -- (11.5, 2.6773) -- cycle;

\draw[->, thick] (3.7, 2.7) -- (4.2, 2.7);

\draw [->] (4.3, -0.2) -- (11, -0.2) node[anchor=west] {$t$};
\filldraw [black] (4.5, -0.2) circle (1.5pt) node[anchor=north] {\small$\tau-3$};
\filldraw [black] (6.5, -0.2) circle (1.5pt) node[anchor=north] {\small$\tau$};
\filldraw [black] (8.5, -0.2) circle (1.5pt) node[anchor=north] {\small$\tau+1$};
\filldraw [black] (10.5, -0.2) circle (1.5pt) node[anchor=north] {\small$\tau + T$};

\node at (8.5, 5.8) {Generation of $13\times3\times60$ context-aware grid};

\fill[green, opacity=0.2] (4.5, 1.03) -- (4.5, 2.1) -- (10.5, 2.1) -- (10.5, 1.03) -- cycle;
\fill[green, opacity=0.2] (10.5, 2.1) -- (10.5, 1.03) -- (12, 1.896) -- (12, 2.946) -- cycle;
\fill[green, opacity=0.2] (12, 2.946) -- (6, 2.946) -- (4.5, 2.1) -- (10.5, 2.1);

\draw[green, opacity=1, dashed] (4.5, 2.1) -- (10.5, 2.1);
\draw[green, opacity=1, dashed] (4.5, 1.03) -- (10.5, 1.03);
\draw[green, opacity=1, dashed] (12, 2.946) -- (6, 2.946);

\node at (13.1, 0.6) {\small$11\times1\times80$};
\node at (15.0, 5.8) {CNN processing};
\node at (13.3, 4.8) {\small$3\times3$ conv};

\draw (12.5, 1) -- (12.5, 4.3) -- (13.5, 4.3) -- (13.5, 1) -- cycle;
\draw (12.5, 4.3) -- (12.933, 4.55) -- (13.933, 4.55) -- (13.5, 4.3);
\draw (13.933, 4.55) -- (13.933, 1.25) -- (13.5, 1);
\draw[dashed] (12.933, 4.55) -- (12.933, 1.25) -- (12.5, 1);
\draw[dashed] (12.933, 1.25) -- (13.933, 1.25);

\foreach \i in {1, 2, 3, 4, 5, 6, 7, 8, 9, 10, 11}
{
    \draw (12.5, 1 + 0.3 * \i) -- (12.933, 1.25 + 0.3 * \i);
    \draw (13.5, 1 + 0.3 * \i) -- (13.933, 1.25 + 0.3 * \i);
}

\fill[green, opacity=0.3] (12.5, 1.9) -- (12.5, 2.2) -- (13.5, 2.2) -- (13.5, 1.9) -- cycle;
\fill[green, opacity=0.3] (12.5, 2.2) -- (12.933, 2.45) -- (13.933, 2.45) -- (13.5, 2.2) -- cycle;
\fill[green, opacity=0.3] (13.933, 2.45) -- (13.933, 2.15) -- (13.5, 1.9) -- (13.5, 2.2) -- cycle;

\draw[green, opacity=0.9, dashed] (12.5, 1.9) -- (13.5, 1.9);
\draw[green, opacity=0.9, dashed] (12.5, 2.2) -- (13.5, 2.2);
\draw[green, opacity=0.9, dashed] (12.933, 2.45) -- (13.933, 2.45);

\draw[opacity=0.7] (10.5, 2.1) -- (12.5, 2.2);
\draw[opacity=0.7] (10.5, 1.03) -- (12.5, 1.9);
\draw[opacity=0.7] (12, 2.946) -- (12.933, 2.45);
\draw[opacity=0.7] (12, 1.896) -- (12.5, 2.04);

\draw (14.5, 1.3) -- (14.5, 4.) -- (15.5, 4.) -- (15.5, 1.3) -- cycle;
\draw (14.5, 4.) -- (14.933, 4.25) -- (15.933, 4.25) -- (15.5, 4.);
\draw (15.933, 4.25) -- (15.933, 1.55) -- (15.5, 1.3);
\draw[dashed] (14.933, 4.25) -- (14.933, 1.55) -- (14.5, 1.3);
\draw[dashed] (14.933, 1.55) -- (15.933, 1.55);

\foreach \i in {1, 2, 3, 4, 5, 6, 7, 8, 9}
{
    \draw (14.5, 1.3 + 0.3 * \i) -- (14.933, 1.55 + 0.3 * \i);
    \draw (15.5, 1.3 + 0.3 * \i) -- (15.933, 1.55 + 0.3 * \i);   
}
\node at (15.1, 1) {\small$9\times12$};
\node at (15.4, 4.6) {\small$3\times1$ conv};

\fill[blue, opacity=0.4] (12.5, 3.1) -- (13.5, 3.1) -- (13.5, 4) -- (12.5, 4) -- cycle;
\fill[blue, opacity=0.4] (12.5, 4) -- (12.933, 4.25) -- (13.933, 4.25) -- (13.5, 4) -- cycle;
\fill[blue, opacity=0.4] (13.5, 4) -- (13.933, 4.25) -- (13.933, 3.35) -- (13.5, 3.1) -- cycle;

\draw[blue, opacity=0.8, dashed] (12.5, 3.1) -- (13.5, 3.1);
\draw[blue, opacity=0.8, dashed] (13.5, 4) -- (12.5, 4);
\draw[blue, opacity=0.8, dashed] (12.933, 4.25) -- (13.933, 4.25);

\fill[blue, opacity=0.4] (14.5, 3.4) -- (14.5, 3.7) -- (15.5, 3.7) -- (15.5, 3.4) -- cycle;
\fill[blue, opacity=0.4] (14.5, 3.7) -- (14.933, 3.95) -- (15.933, 3.95) -- (15.5, 3.7) -- cycle;
\fill[blue, opacity=0.4] (15.5, 3.7) -- (15.933, 3.95) -- (15.933, 3.65) -- (15.5, 3.4) -- cycle;

\draw[blue, opacity=0.8, dashed] (14.5, 3.7) -- (15.5, 3.7);
\draw[blue, opacity=0.8, dashed] (14.5, 3.4) -- (15.5, 3.4);
\draw[blue, opacity=0.8, dashed] (14.933, 3.95) -- (15.933, 3.95);

\draw[opacity=0.7] (13.5, 4) -- (14.5, 3.7);
\draw[opacity=0.7] (13.933, 4.25) -- (14.933, 3.95);
\draw[opacity=0.7] (13.5, 3.1) -- (14.5, 3.4);
\draw[opacity=0.7] (13.933, 3.35) -- (14.5, 3.5);

\draw (16.5, 1.4 + 0.5) -- (16.5, 2.9 + 0.5) -- (17, 2.9 + 0.5) -- (17, 1.4 + 0.5) -- cycle;
\draw (16.5, 2.9 + 0.5) -- (16.933, 3.15 + 0.5) -- (17.433, 3.15 + 0.5) -- (17, 2.9 + 0.5);
\draw (17.433, 3.15 + 0.5) -- (17.433, 1.65 + 0.5) -- (17, 1.4 + 0.5);

\foreach \i in {1, 2, 3, 4, 5}
    \draw (16.5, 1.9 + 0.3 * \i) -- (17, 1.9 + 0.3 * \i) -- (17.433, 2.15 + 0.3 * \i);
\node at (16.8, 1.55) {\small$5\times12$};
\node at (17, 4.3) {\small$2\times1$};
\node at (17, 4.) {\small maxpool};

\draw[->] (17.6, 2.8) -- (17.9, 4.0);
\draw[->] (17.6, 2.8) -- (17.9, 1.6);
\node at (18.8, 6.3) {\small Fully};
\node at (18.8, 6.1) {\small Connected};
\draw[black] (20.0,5.8) rectangle (18,3.50);
\node at (19.0, 5.3) {Lateral};
\node at (19.0, 4.9) {decision};
\node at (19.0, 4.5) {network};
\node at (19.0, 3.9) {80:100:3};
\draw[->] (18.85, 3.5) -- (18.85, 3.15);
\node at (18.9, 2.8) {\large$\left[S, L, R \right]$};
\draw[black] (20.0,2.4) rectangle (18,0.0);
\node at (19.0, 1.8) {Longitudinal};
\node at (19.0, 1.4) {decision};
\node at (19.0, 1.0) {network};
\node at (19.0, 0.5) {80:100:2};
\draw[->] (18.85, 0.0) -- (18.85, -0.35);
\node at (18.9, -0.7) {\large$\left[C, B \right]$};

\draw[dashed, opacity=0.8] (3.9, -0.7) rectangle (17.7, 6.2);

\end{tikzpicture}
\caption{Context-aware grid map and CNN decision engine for AV decision making}
\label{Model}
\end{figure*}
\begin{algorithm}
\begin{algorithmic}[1]
\State {\textbf{Inputs}: The set of past track history $\{\mathbf{X}_h\}_{i=1}^{\mathcal{N}}$ where $\mathcal{N}$ is the number of surrounding vehicles detected by the sensors, pre-trained weights for MNN, current time step $t$}, prediction horizon $T_H$
\State {\textbf{Initialize}: $\mathcal{G}$ is a $13\times3\times60$ array of zeros}
\State {\textbf{Step 1}: Obtain the \textit{context-aware grid $\mathcal{G}$}:}
\For{$v \leftarrow 1$ to $\mathcal{N}$}
\For{$\tau \leftarrow t-30$ to $t+T$}
\If{$\tau \leq$ t}
\State{$\mathcal{G}(i, j, \tau) = 1.0$, where $(i, j)$ is the grid \newline \phantom a \phantom a \phantom a\phantom a indices obtained using the relative position $\mathbf{X}_h$ \newline \phantom a \phantom a \phantom a   \phantom. of vehicle $v$ w.r.t ego vehicle's position at time $\tau$.}
\Else 
\State{$\mathcal{G}(i, j, \tau) = \mathbf{P}(\tau)$, where $(i, j)$ is the grid  \newline \phantom a \phantom a \phantom a \phantom a indices obtained using the predicted position of \newline \phantom a \phantom a \phantom a \phantom . vehicle $v$ (from MNN), and $\mathbf{P}(\tau)$ is calculated \newline \phantom a \phantom a \phantom a \phantom . using Equ.\ref{eq:decay}}
\EndIf
\EndFor
\EndFor
\caption{DST-CAN Context-aware grids generation}
\label{DST-CAN code}
\end{algorithmic}
\end{algorithm}
Based on their past trajectories, POMs are created for the future predicted positions of the surrounding vehicles present at the current instant in the occupancy grid map. The probability that a car will be at the predicted location decreases exponentially with time. The difference in the position of a non-ego vehicle from time step ($t-1$) to $t$ is given as input to the trained MNN. The MNN then outputs the next predicted change in the position. The predicted position can be computed as shown below,
\begin{equation}
    \hat{x}_{t+1} = x_t + \Delta \hat{x}_t;~~
    \hat{y}_{t+1} = y_t + \Delta \hat{y}_t 
    \label{eq:increment}
\end{equation}
The vehicle will be at that predicted position in the grid with a probability $P(t)$ as given in Equ.\ref{eq:decay} and as shown in Fig.\ref{grid_t+1}. Since there is always uncertainty in a vehicle’s future predicted position, it is assumed that the vehicle could be at any of the eight surrounding grids encircled by the MNN predicted grid. It is anticipated that the vehicle could be at any one of the surrounding grids with a probability \((1 - P(t))/8\) as shown in Fig.\ref{grid_t+1} for the first future time step. Once the first predicted location is found for one future instant, the model takes the change in location from the last time step and passes that change to the MNN to get the new predicted location change for the second future step. This new location calculated as a running sum will then have the probability of occupancy calculated according to Equ.\ref{eq:decay} as given by Equ.\ref{eq:increment2} (shown in Fig.\ref{grid_t+k}).
\begin{equation}
\begin{aligned}
    \hat{x}_{t+(k+1)} = \hat{x}_{t+k} + \Delta \hat{x}_{t+k};~~
    \hat{y}_{t+(k+1)} = \hat{y}_{t+k} + \Delta \hat{y}_{t+k}
\end{aligned}
\label{eq:increment2}
\end{equation}
where $k = 1,2,..., (m-1)$. A context-aware grid map is formed in this way, as shown left part of Fig.\ref{Model} for the rest of the prediction horizon. 

\subsubsection{Predictive Manoeuvre Decisions Engine} \label{NN_action_engine}
\begin{figure}[h]
\centering
\begin{adjustbox}{max size={.5\textwidth}{.2\textheight}}
\begin{tikzpicture}
\begin{axis}[
    title={},
    xlabel={Epochs},
    ylabel={Binary cross entropy loss},
    xmin=0, xmax=12,
    ymin=0.16, ymax=0.25,
    xtick distance=1,
    ytick distance=0.01,
    legend pos=north east,
    ymajorgrids=true,
    grid=both,
    grid style=dashed,
]
\addplot[
    color=orange,
    mark=square,
    ]
    coordinates {
    (1,0.186577722430229)(2,0.17488445341587)(3,0.172613397240638)(4,0.170961961150169)(5,0.169855341315269)(6,0.169281303882598)(7,0.168322265148162)(8,0.167983159422874)(9,0.16750918328762)(10,0.167064353823661)(11,0.166933417320251)(12,0.166661918163299)
    };
\addplot[
    color=blue,
    mark=square,
    ]
    coordinates {
    (1,0.244860678911209)(2,0.226984083652496)(3,0.223026424646377)(4,0.221123024821281)(5,0.219469100236892)(6,0.218224376440048)(7, 0.217213824391365)(8,0.216486796736717)(9,0.215716883540153)(10,0.215354710817337)(11,0.214793175458908)(12,0.214501246809959)
    };
\legend{$\text{with data pruning}$, $\text{without data pruning}$}
\end{axis}
\end{tikzpicture}
\end{adjustbox}
\caption{Network training error history}
\label{training_graph}
\end{figure}
The spatio-temporal context and occupancy map information are embedded into context-aware grid maps and are used to determine the motion planning decisions. Here, a CNN is used to extract the features from the context-aware grid maps, and fully connected layers find the relationship between features and decision space. The architecture is shown in the right part of Fig.\ref{Model}. The CNN employs two three-dimensional convolutional layers with leaky-ReLU activation function and one max-pooling layer. The extracted features are first flattened and passed through two fully connected lateral and longitudinal decision networks. Each of these fully connected networks has one hidden layer with $100$ nodes. The last layer employs softmax function and outputs the probability distributions of lateral and longitudinal motion planning decisions. The output with the highest probability is chosen as the final decision taken by the ego vehicle.
\begin{algorithm}
\begin{algorithmic}[1]
\State {\textbf{Input}: Context-Aware Spatio Temporal Grids for every instant as tensors with manoeuvre decisions (either Ground Truth (GT) decisions or human decisions) at that moment}
\State {\textbf{Initialize}: Convolutional Neural Network (CNN) with final output dimension of manoeuvre action classes and a blank list [L] for storing tensor indexes}
\State {\textbf{Step 1:} For first epoch, pass all tensors through CNN}
\For{$i \leftarrow 1$ to Total number of social tensors}
\If{predicted decision = actual decision (either GT\phantom a  \quad decision or human decision)}
\State{Save the tensor index $i$ to the list [L]}
\EndIf
\EndFor
\State{\textbf{Step 2:} Remove the tensors in list L from training dataset and get a new training dataset with remaining data}
\State{\textbf{Step 3:} Train the CNN with new training dataset}
\For{$i \leftarrow 1$ to E number of epochs}
\For{$i \leftarrow 1$ to batch size B}
\State{predict the maneuver action classes as probabili \phantom a \phantom . \quad \phantom a  -ties and calculate the Binary Cross Entropy (BCE) \phantom a \phantom a \quad loss $\mathcal{L}$ as given in Equ.\ref{eq:BCEloss}}
\EndFor
\EndFor
\caption{DST-CAN data pruning and training pseudo code}
\label{DST-CAN training code}
\end{algorithmic}
\end{algorithm}

\subsubsection{Imitation Learning for DST-CAN}
DST-CAN is trained using imitation learning \cite{Bojarski_2016} approach, where the network tries to mimic the expert behaviour in different conditions. For PMP, imitation learning from real-world traffic data has been utilized to generate the decision class. Here, human decisions in real traffic and traffic rule-based ground truth decisions are used for imitation learning. The binary cross-entropy loss given in Equ.\ref{eq:BCEloss} is minimized to generate an update rule. 
\begin{equation}
    \mathcal{L} = - \frac{1}{N}\sum_{i=1}^{N}y_ilog(p(y_i))+(1-y_i)log(1-p(y_i))
    \label{eq:BCEloss}
\end{equation}
where $y_i$ is the ground truth and $p(y_i)$ is the predicted output. $N$ is the total number of samples. The model is implemented in PyTorch and is trained using an RMSProp optimizer.
Since the dataset is a highly imbalanced long-tail dataset (sample imbalance described in the supplementary material), a data-pruning method has been applied to remove the bias from the training dataset. The data samples that give good classification scores for the first epoch are removed from the training dataset. The remaining data samples are used for further training the network. The training loss convergence graph for a $3$-sec prediction horizon with and without the data pruning method (as described in Algorithm \ref{DST-CAN training code}) is given in Fig.\ref{training_graph}. This work uses an Intel $i5$ processor with $32$ GB RAM and an NVIDIA GeForce RTX $2080$ Ti graphic card to generate context-aware grid maps from trajectory data and process those maps with CNN and fully connected layers in only $0.073$ seconds.
\section{Performance Evaluation of DST-CAN}
\label{Performance Evaluation of DST-CAN}
In this section, the performance evaluation results of DST-CAN using the benchmark traffic datasets are presented. First, we present details of the data sets used in this study and next the simulation setup for these studies are described, followed by the performance analysis of DST-CAN for different prediction horizons. The analysis of conflicting cases and study on near-collision scenarios are done. Finally, the performance of DST-CAN using a $3$-sec prediction horizon is compared with the earlier state-of-the-art approach of CS-LSTM \cite{deo2018convolutional} in terms of the number of correct decisions made for safe and efficient driving by an AV.

\subsection{Traffic Dataset Description} 
\begin{table}[]
\centering
\begin{tabular}{|l|l|l|l|l|l|}
\hline
          & 1 sec         & 2 sec         & 3 sec         & 4 sec         & 5 sec         \\ \hline
CV        & 0.73          & 1.78          & 3.13          & 4.78          & 6.68          \\ \hline
CV-GMM \cite{deo2018would}   & 0.66          & 1.56          & 2.75          & 4.24          & 5.99          \\ \hline
GAIL-GRU \cite{kuefler2017imitating} & 0.69          & 1.56          & 2.75          & 4.24          & 5.99          \\ \hline
LSTM      & 0.68          & 1.65          & 2.91          & 4.46          & 6.27          \\ \hline
MATF \cite{Zhao2019MultiTensorFusion}     & 0.67          & 1.51          & 2.51          & 3.71          & 5.12          \\ \hline
CS-LSTM \cite{deo2018convolutional}  & 0.61          & 1.27          & 2.09          & 3.10          & 4.37          \\ \hline
S-LSTM \cite{alahi2016social}   & 0.65          & 1.31          & 2.16          & 3.25          & 4.55          \\ \hline
M-LSTM \cite{deo2018multi}   & 0.58          & 1.26          & 2.12          & 3.24          & 4.66          \\ \hline
UADMF \cite{tang2022prediction}    & 0.58          & 1.24          & 2.06          & 3.09          & 4.38          \\ \hline
UST \cite{he2020ust}      & 0.58          & 1.20          & 1.96          & 2.92          & 4.12          \\ \hline
UST-180 \cite{he2020ust}  & 0.56          & 1.15          & 1.82          & 2.58          & 3.45          \\ \hline
MFP \cite{Tang2019MFP}      & 0.54          & 1.16          & 1.9          & 2.78          & 3.83          \\ \hline
SAMMP $(k=1)$ \cite{mercat2020multi}    & 0.51          & 1.13          & 1.88          & 2.81          & 3.67          \\ \hline
STDAN \cite{chen2022intention}    & 0.42          & 1.01          & 1.69          & 2.56          & 4.05          \\ \hline
GRIP \cite{li2019Grip}     & 0.37          & 0.86          & 1.45          & 2.21          & 3.16          \\ \hline
GRIP++ \cite{li2019Grip++}   & 0.38          & 0.89          & 1.45          & 2.14          & 2.94          \\ \hline
STA-LSTM \cite{Lin2022STA-LSTM} & 0.37          & 0.98          & 1.71          & 2.63          & 3.78          \\ \hline
DeepTrack \cite{katariya2022Deep_track} & 0.47          & 1.08          & 1.83          & 2.75          & 3.89          \\ \hline
MMnTP $(k=1)$ \cite{mozaffari2023multimodal} & \textbf{0.36}          & 0.96          & 1.69          & 2.56          & 3.55          \\ \hline
DST-CAN   & \textbf{0.36} & \textbf{0.85} & \textbf{1.38} & \textbf{1.92} & \textbf{2.74} \\ \hline
\end{tabular}
\caption{Trajectory prediction results compared with state-of-the-art unimodal approaches. The evaluation is based on Root Mean Square Error (RMSE) values, measured in meters.}
\label{prediction_summary}
\end{table}

\begin{table*}
\centering
  \begin{tabular}{lSSSSSS}
    \toprule
    \multirow{2}{*}{Prediction horizon} &
      \multicolumn{1}{c}{Lateral Consensus cases} &
      \multicolumn{1}{c}{Lateral Conflict cases} &
      \multicolumn{1}{c}{Longitudinal Consensus cases} &
      \multicolumn{1}{c}{Longitudinal Conflict cases} \\
      & {accuracy (\%)}& {accuracy (\%)}& {accuracy (\%)} & {accuracy (\%)}\\
      \midrule
    1 sec & 99.62 & 13.79 & 95.14 & 32.31 \\ \hline
    3 sec & \textbf{99.67} & \textbf{13.85} & \textbf{95.35} & \textbf{32.81}\\ \hline
    5 sec & 99.64 & 13.61 & 95.24 & 32.35 \\
    \bottomrule
  \end{tabular}
  \caption{Performance analysis for different time horizons}
  \label{ablation_table}
\end{table*}
The traffic datasets NGSIM I-80 and US-101 \cite{colyar2007us101} have been used in this work. The two datasets consist of $45$ minutes of traffic data at two different locations. Each dataset has three segments of low, medium, and highly congested traffic data. Each segment has $15$ minutes of traffic data. The datasets have data sampled at $10Hz$. The raw dataset gives the vehicles' positions and vehicle IDs at each time step. As a first step, the dataset has to be pre-processed for prediction purposes. Each vehicle in the time step is represented as an ego vehicle. A vehicle takes from $5$ to $10$ seconds to change the lane as described in \cite{Song2022LaneChangeReview}. It has been assumed that a vehicle lane change manoeuvre could take $8$ seconds. This work assumes that if a vehicle's lane ID changed from the previous $4$ seconds to the next $4$ seconds, the vehicle changed its lane. From this change in the lane, the lateral manoeuvre is recorded. Suppose the lane number decreases within this $\pm$ $4$ seconds, then the vehicle changes to the left lane. Also, a braking decision is identified if the vehicle's average speed over the next $5$ seconds is lower than $0.8$ times the current speed.

It has also been found that there are scenarios where the actual decisions taken by a human driver are not safe. Hence, if only human decisions are used for learning the possible driving decisions, then the AV can also make unsafe decisions. Therefore, there is a need for generating traffic rule-based decisions for each scenario to provide for safe actions. For this reason, Ground Truth (GT) using a rule-based approach are generated for both the datasets. More details about GT can be found in the section \ref{Simulation setup} and also in the supplementary material.

\subsection{Simulation Setup} \label{Simulation setup} The NGSIM dataset comprises many scenarios in the study regions that include both on-ramps and off-ramps. These different scenarios include significant changes in traffic patterns. Hence, these datasets have been used to study the performance of the proposed DST-CAN approach. The low and highly-congested traffic data from both the datasets have been used for training and the medium-congested data for testing. A summary of the details of the datasets is given in the supplementary materials. For constructing the spatio-temporal grids, the considered sensor range in an AV is $90$ feet. Each processed occupancy map consists of vehicles in the range of $\pm 90$ feet and two adjacent lanes of the ego vehicle. It is assumed that each lane is $15$ feet wide. For prediction, $3$ sec of past data is used. Three prediction time horizons ($1$, $3$ and $5$ sec) were used to find the one that gave a better result. Most of the data in the datasets correspond to decision-making during cruising in the same lane. Since keeping the same lane manoeuvre is a major portion of the dataset, only a small portion ($20\%$) of those manoeuvres are taken for imitative driving model training. This processing allows the model to predict lane-changing and longitudinal manoeuvres well.

Here, we present a few examples of traffic rule-based ground truth generation for training the network.  a) If the distance between the AV and the vehicle in the front is less than two grids, then one should explore possible lane change manoeuvre. If at least two grids (i.e., $\sqrt{5}$ in L2 distance from the ego vehicle) are unoccupied on the adjacent lanes on both the forward and rear sides, it is safe to switch the lane; b) If there is more than two grid space in front of the ego vehicle and the velocity of the front vehicle is same or more than the ego vehicle, then it is safe to cruise in the same lane; c) If none of the adjacent lanes is free to move safely, then the ego vehicle should brake on the same lane. More details on traffic rule-based GT generation and rules can be found in the supplementary materials. Two DST-CAN models, DST-CAN trained with actual human decision data and rule-based GT data are used to understand the importance of following standard traffic rules.

\subsection{Case Studies}
The trained DST-CAN model has been evaluated using the testing dataset with moderate congestion traffic. We have conducted four different case studies namely, a) trajectory prediction comparison with baseline models, b) the effect of prediction horizon, c) the effect of human bias and d) a study on near-collision scenarios.

\subsubsection{Trajectory prediction comparison with baseline models}
We have presented these trajectory prediction results of DST-CAN for horizons $(1s$ to $5s)$ and compared the results with CV and state-of-the-art unimodal deep neural network. The results are reported in Table \ref{prediction_summary}. From the table, we can see that DST-CAN is better compared to other algorithms for higher prediction horizons and comparable with MMnTP $(k=1)$ for $1s$ horizon.

\subsubsection{Study on effect on the prediction horizon}
Typically, a lane change manoeuvre in AV requires a time duration between $4$ to $8$ seconds \cite{Song2022LaneChangeReview}, and past $3$ seconds trajectory data are used for predicting the future trajectory. Hence, different prediction horizons ($1$, $3$, $5$ seconds) have been used to study the effect of prediction horizon on the performance of DST-CAN. Table \ref{prediction_summary} shows the trajectory prediction results of DST-CAN with other recent unimodal prediction models given in MMnTP \cite{mozaffari2023multimodal} and UADMF \cite{tang2022prediction}. Table \ref{ablation_table} shows the comparison results for different horizons. In order to study the performance, we divide data samples into two categories: consensus and conflict cases. Consensus cases are those where actual human decisions are the same as the GT decisions. Conflict cases are those where human decisions and GT decisions are different. In Table \ref{ablation_table}, the DST-CAN model is trained based only on human decisions. The accuracies for both the consensus and conflict cases are shown in Table \ref{ablation_table}. A prediction horizon of one second falls short in accurately understanding the intentions of a manoeuvre. In the case of predictions that span $5$ seconds, most vehicles exit the ego vehicle's context region. This leads to Probabilistic Occupancy Maps (POMs) predicted after $4.2$ seconds showing zero occupancy, influencing manoeuvre decisions. The table reveals that a prediction horizon of $3$ seconds is more effective and has been utilized for the remainder of the investigation. 

\subsubsection{Effect of human bias}
\begin{figure*}[h]
\centering
\begin{subfigure}[b]{0.24\textwidth}
\includegraphics[width=\textwidth]{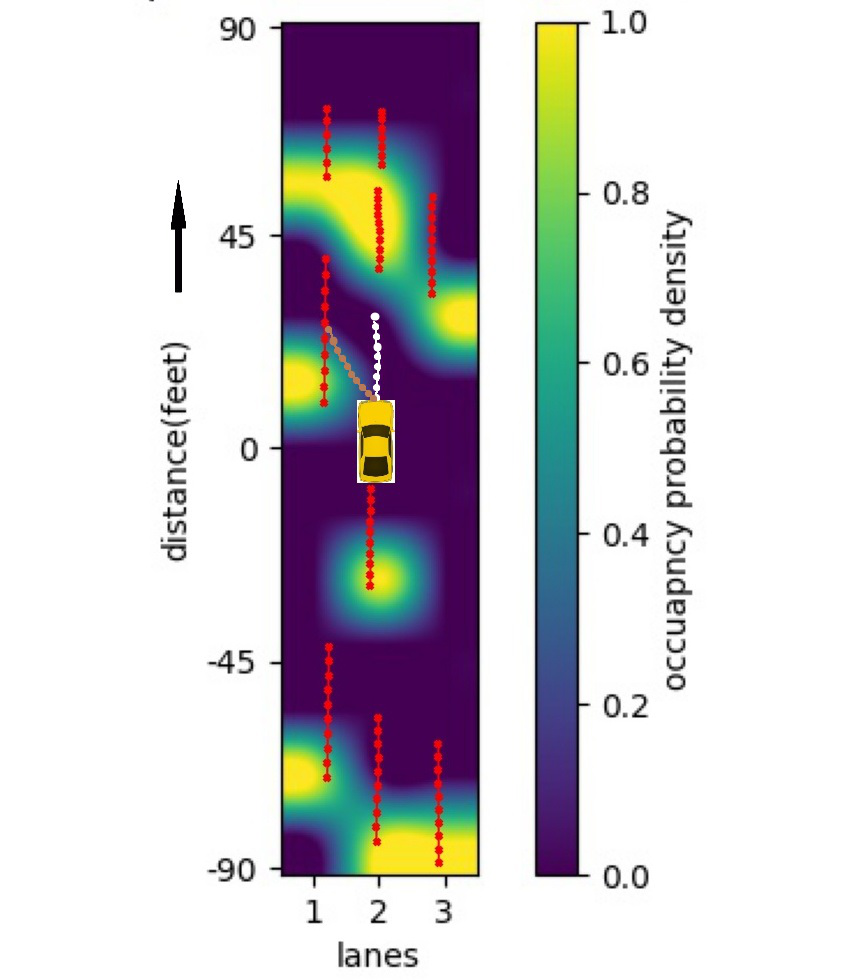}
\caption{Safety scenario}
\label{tensor224130}
\end{subfigure}
\begin{subfigure}[b]{0.24\textwidth}
\includegraphics[width=\textwidth]{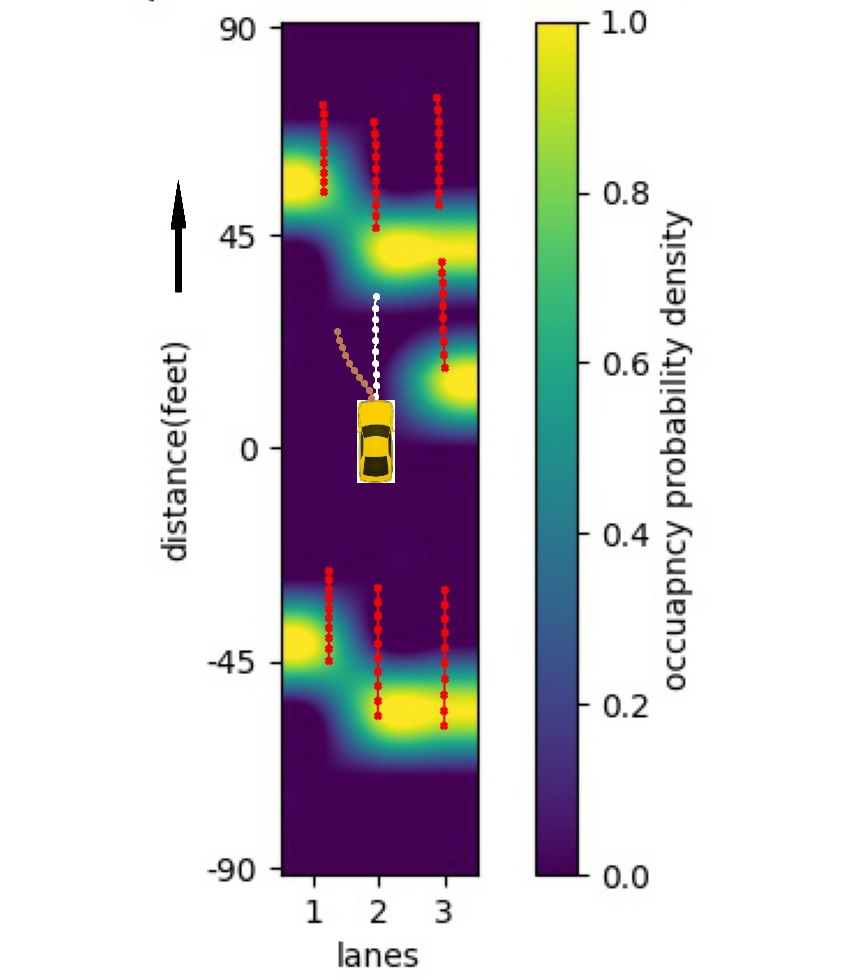}
\caption{Efficiency scenario}
\label{tensor361452}
\end{subfigure}
\begin{subfigure}[b]{0.24\textwidth}
\includegraphics[width=\textwidth]{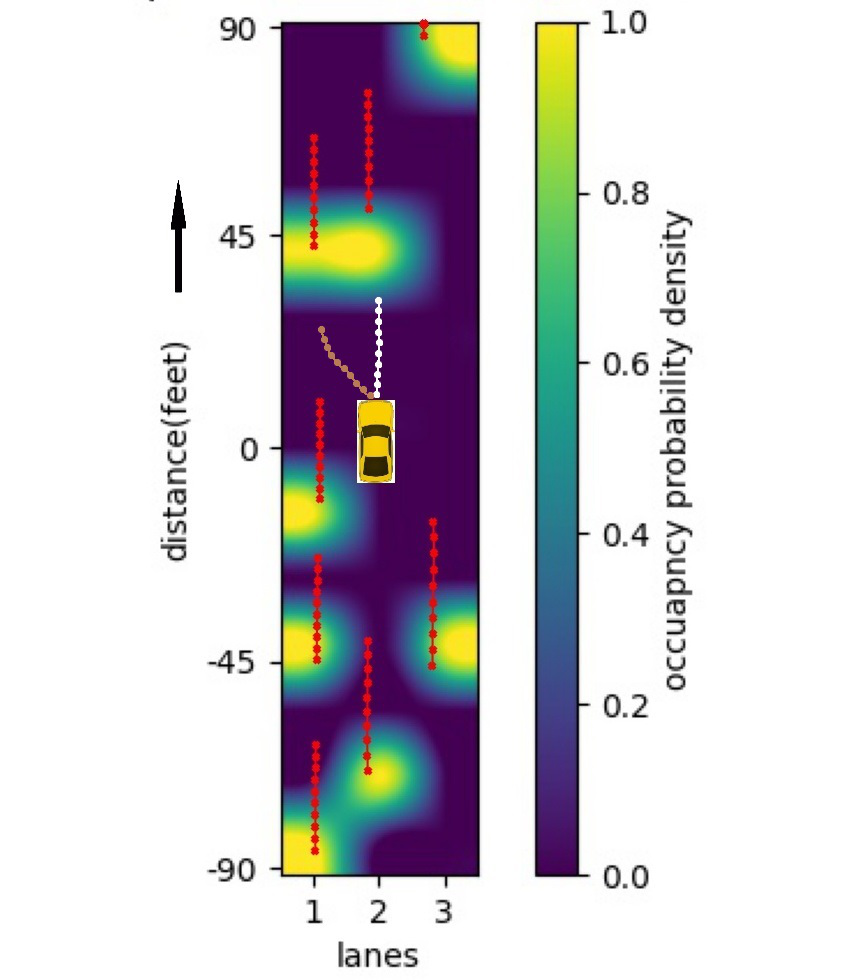}
\caption{Safety \& efficiency scenario}
\label{tensor8563}
\end{subfigure}
\begin{subfigure}[b]{0.24\textwidth}
\includegraphics[width=\textwidth]{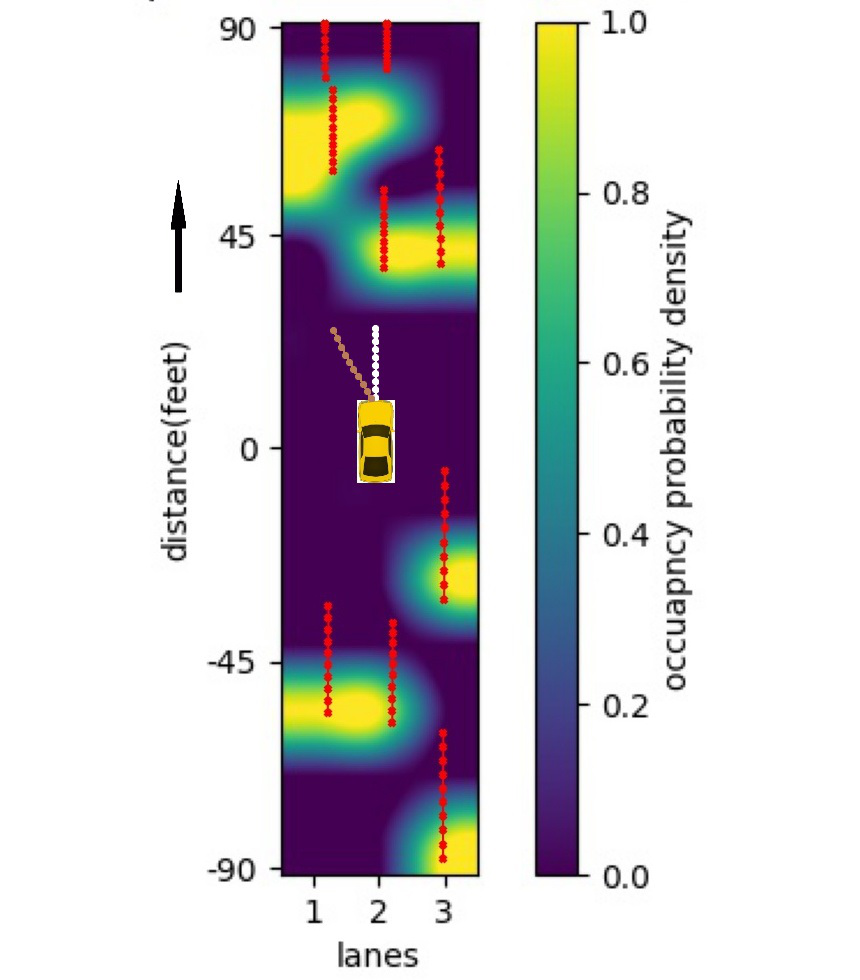}
\caption{Low efficiency scenario}
\label{tensor21708}
\end{subfigure}
\caption{Occupancy heat maps of different prediction scenarios (DST-CAN decisions in white dots and human decisions in brown dots)}
\label{heatmaps}
\end{figure*}
The human decision depends on individual skills, and the decisions may differ for different drivers, whereas the traffic rule-based decision is independent of human bias. In order to study the effect of this human bias, we have selected the samples where the rule-based decision is different from the human decision as conflicting cases. Here, DST-CAN trained using rule-based is compared with the human decision.
\begin{table}[h!]
\centering
\begin{tabular}{|c|c|c|}
\hline
\begin{tabular}[c]{@{}c@{}}Near collision\\ scenarios\end{tabular} &
  \begin{tabular}[c]{@{}c@{}}I80\\ testing dataset\end{tabular} &
  \begin{tabular}[c]{@{}c@{}}US101\\ testing dataset\end{tabular} \\ \hline
\begin{tabular}[c]{@{}c@{}}DST-CAN\\ performance\end{tabular} & 8.028 \% & 9.989 \%  \\ \hline
\begin{tabular}[c]{@{}c@{}}Traffic\\ observations\end{tabular}        & 8.163 \% & 10.037 \% \\ \hline
\end{tabular}
\caption{Performance analysis for near collision scenarios. The results indicate the percentage of instances in which surrounding vehicles enter the near-collision region around the ego vehicle.}
\label{Tab_near_collision_scenarios}
\end{table}

In these cases, the DST-CAN predicted better decisions by correctly understanding the behaviours of the surrounding vehicles from their past and the predicted future trajectories. Some of these conditions for mismatching manoeuvres are shown in Fig.\ref{heatmaps}. This figure shows safety, efficiency, safety and efficiency and low efficiency scenarios. Safety and efficiency are defined based on the decision taken by the ego vehicle for a given situation. 
In each scenario, the figure provides surrounding vehicles and their trajectories. The red dots show the trajectories of surrounding vehicles for the next one second. Only the next one-second of future trajectories (as recorded in the dataset) has been shown for visualisation purposes. Each surrounding vehicle has ten dots because trajectories are recorded at 10 Hz. The predicted heatmaps of the surrounding vehicles for the immediate next step are shown in the same figure. The position of the ego vehicle is represented by the yellow car in the image. The images in Fig.\ref{heatmaps} show the predicted occupancy heat map and actual trajectories of the surrounding vehicles with the ego vehicle at the centre. The ego vehicle trajectory followed by the human decision is shown in brown dots, and the DST-CAN decision is shown in white dots.

In Fig.\ref{tensor224130}, it can be seen that there is a vehicle very near to the ego vehicle on the left side. The human driver decided to brake and take the left lane. From the future trajectory of that nearby vehicle (as can be seen from the red dots), it is clear that if the ego vehicle tries to take a left, there is a high chance of a near-collision scenario (within $15$ ft of the ego vehicle). In this scenario, the DST-CAN decides to cruise on the same lane without the risk of a near-collision scenario. Hence, in this case, the human decision is \textit{unsafe}, and DST-CAN predicts a much \textit{safer} decision of cruising on the same lane. In another scenario (as shown in Fig.\ref{tensor361452}), it can be seen that the road in front of the ego vehicle is free (front vehicle $45$ ft away). Even though the lane is free of nearby vehicles, the human driver decides to change to the left lane. Since there is no vehicle in the left lane no near collision alert; however, changing lanes needs additional effort. Since there is no need for a lane change, DST-CAN's decision was to remain in the same lane. This clearly shows that DST-CAN is \textit{efficient} and \textit{safe} for manoeuvre planning. For the scenario shown in Fig.\ref{tensor8563}, moving to the left lane is not very safe as vehicles are in the rear (very nearby) and forward end (within $40$ ft) of the left lane. In this scenario, following the same lane is \textit{much safer and more efficient}. For example, in Fig.\ref{tensor21708}, the human decision is to change lanes on the left side and cruise, which is a safe decision; however, this is inefficient. DST-CAN decided to slow down in the same lane. However, one can see from the heat map that cruising on the same lane is the most efficient manoeuvre. It is important to note that the DST-CAN model is trained on a large dataset of real-world driving scenarios, and it may sometimes make decisions that are not optimal in terms of efficiency but are necessary for safety. If the map of the environment is added along with the dataset to the DST-CAN model, one can avoid low-efficiency scenarios. 
\begin{table*}[h!]
\centering
  \begin{tabular}{lSSSSSSSS}
    \toprule
    \multirow{2}{*}{Training data used} &
      \multicolumn{2}{c}{Lateral Consensus cases} &
      \multicolumn{2}{c}{Lateral Conflict cases} &
      \multicolumn{2}{c}{Longitudinal Consensus cases} &
      \multicolumn{2}{c}{Longitudinal Conflict cases} \\
      & {DST-CAN} & {CS-LSTM} & {DST-CAN} & {CS-LSTM} & {DST-CAN} & {CS-LSTM} & {DST-CAN} & {CS-LSTM} \\
      \midrule
    Human decision data & \textbf{99.67 \%} & 99.45 \% & \textbf{13.85 \%} & 8.79 \% & \textbf{95.35 \%} & 94.72 \%& \textbf{32.81 \%} & 26.44 \%\\ \hline
    Rule-based GT data & \textbf{99.69 \%} & 98.11 \% & \textbf{99.82 \%} & 78.67 \% & \textbf{99.92 \%} & 97.95 \% & \textbf{99.95 \%} & 87.83 \% \\
    \bottomrule
  \end{tabular}
  \caption{Percentage accuracy comparison for Deep Spatio Temporal Context-Aware decision Network vs. CS-LSTM}
  \label{comparison_table}
\end{table*}

\begin{table*}[h!]
\centering
\begin{tabular}{ccccccccccccc}
\cline{1-6} \cline{8-13}
\multicolumn{1}{|c|}{} &
  \multicolumn{5}{c|}{\textbf{CS-LSTM decision}} &
  \multicolumn{1}{c|}{} &
  \multicolumn{1}{c|}{} &
  \multicolumn{5}{c|}{\textbf{DST-CAN decision}} \\ \cline{1-6} \cline{8-13} 
\multicolumn{1}{|c|}{\textbf{GT decision}} &
  \multicolumn{1}{c|}{\textbf{Same lane}} &
  \multicolumn{1}{c|}{\textbf{\begin{tabular}[c]{@{}c@{}}Take \\ left\end{tabular}}} &
  \multicolumn{1}{c|}{\textbf{\begin{tabular}[c]{@{}c@{}}Take \\ right\end{tabular}}} &
  \multicolumn{1}{c|}{\textbf{Cruise}} &
  \multicolumn{1}{c|}{\textbf{Brake}} &
  \multicolumn{1}{c|}{} &
  \multicolumn{1}{c|}{\textbf{GT decision}} &
  \multicolumn{1}{c|}{\textbf{Same lane}} &
  \multicolumn{1}{c|}{\textbf{\begin{tabular}[c]{@{}c@{}}Take \\ left\end{tabular}}} &
  \multicolumn{1}{c|}{\textbf{\begin{tabular}[c]{@{}c@{}}Take \\ right\end{tabular}}} &
  \multicolumn{1}{c|}{\textbf{Cruise}} &
  \multicolumn{1}{c|}{\textbf{Brake}} \\ \cline{1-6} \cline{8-13} 
\multicolumn{1}{|c|}{\textbf{Same lane}} &
  \multicolumn{1}{c|}{86679} &
  \multicolumn{1}{c|}{616} &
  \multicolumn{1}{c|}{3145} &
  \multicolumn{1}{c|}{0} &
  \multicolumn{1}{c|}{0} &
  \multicolumn{1}{c|}{} &
  \multicolumn{1}{c|}{\textbf{Same lane}} &
  \multicolumn{1}{c|}{89604} &
  \multicolumn{1}{c|}{582} &
  \multicolumn{1}{c|}{254} &
  \multicolumn{1}{c|}{0} &
  \multicolumn{1}{c|}{0} \\ \cline{1-6} \cline{8-13} 
\multicolumn{1}{|c|}{\textbf{\begin{tabular}[c]{@{}c@{}}Take left\end{tabular}}} &
  \multicolumn{1}{c|}{46952} &
  \multicolumn{1}{c|}{104501} &
  \multicolumn{1}{c|}{3733} &
  \multicolumn{1}{c|}{0} &
  \multicolumn{1}{c|}{0} &
  \multicolumn{1}{c|}{} &
  \multicolumn{1}{c|}{\textbf{\begin{tabular}[c]{@{}c@{}}Take left\end{tabular}}} &
  \multicolumn{1}{c|}{1} &
  \multicolumn{1}{c|}{155185} &
  \multicolumn{1}{c|}{0} &
  \multicolumn{1}{c|}{0} &
  \multicolumn{1}{c|}{0} \\ \cline{1-6} \cline{8-13} 
\multicolumn{1}{|c|}{\textbf{\begin{tabular}[c]{@{}c@{}}Take right\end{tabular}}} &
  \multicolumn{1}{c|}{47648} &
  \multicolumn{1}{c|}{858} &
  \multicolumn{1}{c|}{188516} &
  \multicolumn{1}{c|}{0} &
  \multicolumn{1}{c|}{0} &
  \multicolumn{1}{c|}{} &
  \multicolumn{1}{c|}{\textbf{\begin{tabular}[c]{@{}c@{}}Take right\end{tabular}}} &
  \multicolumn{1}{c|}{3} &
  \multicolumn{1}{c|}{1} &
  \multicolumn{1}{c|}{237018} &
  \multicolumn{1}{c|}{0} &
  \multicolumn{1}{c|}{0} \\ \cline{1-6} \cline{8-13} 
\multicolumn{1}{|c|}{\textbf{Cruise}} &
  \multicolumn{1}{c|}{0} &
  \multicolumn{1}{c|}{0} &
  \multicolumn{1}{c|}{0} &
  \multicolumn{1}{c|}{303517} &
  \multicolumn{1}{c|}{7287} &
  \multicolumn{1}{c|}{} &
  \multicolumn{1}{c|}{\textbf{Cruise}} &
  \multicolumn{1}{c|}{0} &
  \multicolumn{1}{c|}{0} &
  \multicolumn{1}{c|}{0} &
  \multicolumn{1}{c|}{310698} &
  \multicolumn{1}{c|}{106} \\ \cline{1-6} \cline{8-13} 
\multicolumn{1}{|c|}{\textbf{Brake}} &
  \multicolumn{1}{c|}{0} &
  \multicolumn{1}{c|}{0} &
  \multicolumn{1}{c|}{0} &
  \multicolumn{1}{c|}{74841} &
  \multicolumn{1}{c|}{289404} &
  \multicolumn{1}{l|}{} &
  \multicolumn{1}{c|}{\textbf{Brake}} &
  \multicolumn{1}{c|}{0} &
  \multicolumn{1}{c|}{0} &
  \multicolumn{1}{c|}{0} &
  \multicolumn{1}{c|}{664} &
  \multicolumn{1}{l|}{363581} \\ \cline{1-6} \cline{8-13} 
\multicolumn{1}{l}{} &
  \multicolumn{1}{l}{} &
  \multicolumn{1}{l}{} &
  \multicolumn{1}{l}{} &
  \multicolumn{1}{l}{} &
  \multicolumn{1}{l}{} &
  \multicolumn{1}{l}{} &
  \multicolumn{1}{l}{} &
  \multicolumn{1}{l}{} &
  \multicolumn{1}{l}{} &
  \multicolumn{1}{l}{} &
  \multicolumn{1}{l}{} &
  \multicolumn{1}{l}{} \\
\multicolumn{1}{l}{} &
  \multicolumn{5}{c}{a) Confusion matrix for CS-LSTM} &
  \multicolumn{1}{l}{} &
  \multicolumn{1}{l}{} &
  \multicolumn{5}{c}{b) Confusion matrix for DST-CAN}
\end{tabular}
\caption{Performance comparison based on confusion matrix for conflict scenarios for DST-CAN and CS-LSTM}
\label{tab:conflict_confusion_mat}
\end{table*}

\begin{table*}[h!]
\centering
\begin{tabular}{ccccccccccccc}
\cline{1-6} \cline{8-13}
\multicolumn{1}{|c|}{} &
  \multicolumn{5}{c|}{\textbf{CS-LSTM decision}} &
  \multicolumn{1}{c|}{} &
  \multicolumn{1}{c|}{} &
  \multicolumn{5}{c|}{\textbf{DST-CAN decision}} \\ \cline{1-6} \cline{8-13} 
\multicolumn{1}{|c|}{\textbf{GT decision}} &
  \multicolumn{1}{c|}{\textbf{Same lane}} &
  \multicolumn{1}{c|}{\textbf{\begin{tabular}[c]{@{}c@{}}Take \\ left\end{tabular}}} &
  \multicolumn{1}{c|}{\textbf{\begin{tabular}[c]{@{}c@{}}Take \\ right\end{tabular}}} &
  \multicolumn{1}{c|}{\textbf{Cruise}} &
  \multicolumn{1}{c|}{\textbf{Brake}} &
  \multicolumn{1}{c|}{} &
  \multicolumn{1}{c|}{\textbf{GT decision}} &
  \multicolumn{1}{c|}{\textbf{Same lane}} &
  \multicolumn{1}{c|}{\textbf{\begin{tabular}[c]{@{}c@{}}Take \\ left\end{tabular}}} &
  \multicolumn{1}{c|}{\textbf{\begin{tabular}[c]{@{}c@{}}Take \\ right\end{tabular}}} &
  \multicolumn{1}{c|}{\textbf{Cruise}} &
  \multicolumn{1}{c|}{\textbf{Brake}} \\ \cline{1-6} \cline{8-13} 
\multicolumn{1}{|c|}{\textbf{Same lane}} &
  \multicolumn{1}{c|}{2405634} &
  \multicolumn{1}{c|}{13781} &
  \multicolumn{1}{c|}{30158} &
  \multicolumn{1}{c|}{0} &
  \multicolumn{1}{c|}{0} &
  \multicolumn{1}{c|}{} &
  \multicolumn{1}{c|}{\textbf{Same lane}} &
  \multicolumn{1}{c|}{2443422} &
  \multicolumn{1}{c|}{5059} &
  \multicolumn{1}{c|}{1092} &
  \multicolumn{1}{c|}{0} &
  \multicolumn{1}{c|}{0} \\ \cline{1-6} \cline{8-13} 
\multicolumn{1}{|c|}{\textbf{\begin{tabular}[c]{@{}c@{}}Take left\end{tabular}}} &
  \multicolumn{1}{c|}{1149} &
  \multicolumn{1}{c|}{2851} &
  \multicolumn{1}{c|}{93} &
  \multicolumn{1}{c|}{0} &
  \multicolumn{1}{c|}{0} &
  \multicolumn{1}{c|}{} &
  \multicolumn{1}{c|}{\textbf{\begin{tabular}[c]{@{}c@{}}Take left\end{tabular}}} &
  \multicolumn{1}{c|}{0} &
  \multicolumn{1}{c|}{4093} &
  \multicolumn{1}{c|}{0} &
  \multicolumn{1}{c|}{0} &
  \multicolumn{1}{c|}{0} \\ \cline{1-6} \cline{8-13} 
\multicolumn{1}{|c|}{\textbf{\begin{tabular}[c]{@{}c@{}}Take right\end{tabular}}} &
  \multicolumn{1}{c|}{1419} &
  \multicolumn{1}{c|}{7} &
  \multicolumn{1}{c|}{2838} &
  \multicolumn{1}{c|}{0} &
  \multicolumn{1}{c|}{0} &
  \multicolumn{1}{c|}{} &
  \multicolumn{1}{c|}{\textbf{\begin{tabular}[c]{@{}c@{}}Take right\end{tabular}}} &
  \multicolumn{1}{c|}{1} &
  \multicolumn{1}{c|}{0} &
  \multicolumn{1}{c|}{4263} &
  \multicolumn{1}{c|}{0} &
  \multicolumn{1}{c|}{0} \\ \cline{1-6} \cline{8-13} 
\multicolumn{1}{|c|}{\textbf{Cruise}} &
  \multicolumn{1}{c|}{0} &
  \multicolumn{1}{c|}{0} &
  \multicolumn{1}{c|}{0} &
  \multicolumn{1}{c|}{2038301} &
  \multicolumn{1}{c|}{14878} &
  \multicolumn{1}{c|}{} &
  \multicolumn{1}{c|}{\textbf{Cruise}} &
  \multicolumn{1}{c|}{0} &
  \multicolumn{1}{c|}{0} &
  \multicolumn{1}{c|}{0} &
  \multicolumn{1}{c|}{2052489} &
  \multicolumn{1}{c|}{690} \\ \cline{1-6} \cline{8-13} 
\multicolumn{1}{|c|}{\textbf{Brake}} &
  \multicolumn{1}{c|}{0} &
  \multicolumn{1}{c|}{0} &
  \multicolumn{1}{c|}{0} &
  \multicolumn{1}{c|}{31570} &
  \multicolumn{1}{c|}{180780} &
  \multicolumn{1}{l|}{} &
  \multicolumn{1}{c|}{\textbf{Brake}} &
  \multicolumn{1}{c|}{0} &
  \multicolumn{1}{c|}{0} &
  \multicolumn{1}{c|}{0} &
  \multicolumn{1}{c|}{271} &
  \multicolumn{1}{l|}{212079} \\ \cline{1-6} \cline{8-13} 
\multicolumn{1}{l}{} &
  \multicolumn{1}{l}{} &
  \multicolumn{1}{l}{} &
  \multicolumn{1}{l}{} &
  \multicolumn{1}{l}{} &
  \multicolumn{1}{l}{} &
  \multicolumn{1}{l}{} &
  \multicolumn{1}{l}{} &
  \multicolumn{1}{l}{} &
  \multicolumn{1}{l}{} &
  \multicolumn{1}{l}{} &
  \multicolumn{1}{l}{} &
  \multicolumn{1}{l}{} \\
\multicolumn{1}{l}{} &
  \multicolumn{5}{c}{a) Confusion matrix for CS-LSTM} &
  \multicolumn{1}{l}{} &
  \multicolumn{1}{l}{} &
  \multicolumn{5}{c}{b) Confusion matrix for DST-CAN}
\end{tabular}
\caption{Performance comparison based on confusion matrix for consensus scenarios for DST-CAN and CS-LSTM}
\label{tab:consensus_confusion_mat}
\end{table*}

\subsubsection{Study for near-collision scenarios}
A study was conducted to evaluate the robustness of the trained DST-CAN model to near-collision scenarios in unknown environments. A region of $15 ft \times 15 ft$ around the ego vehicle was defined to identify surrounding vehicles that are at risk of collision. This region was defined based on the average distance between different vehicles. Any scenario in which a surrounding vehicle entered this region was marked as a near-collision scenario. The percentage of scenarios in the I$80$ and US$101$ datasets where any one surrounding vehicle was inside the ego vehicle's near-collision region is shown in Table \ref{Tab_near_collision_scenarios}. As can be seen from the table, only $8.028 \%$ and $9.989 \%$ of scenarios in the I$80$ and US$101$ datasets, respectively, were near-collision scenarios. This suggests that the decisions made by the DST-CAN model are robust to near-collision scenarios.

\subsection{Performance Comparison}
In this section, we compare the performance of DST-CAN trained using both human decisions and ground truth decisions with the other state-of-the-art approach in the literature \cite{deo2018convolutional}. The past $3$ second of trajectory history in the sensing region is used by DST-CAN and CS-LSTM to predict the manoeuvre decision. For the implementation of \cite{deo2018convolutional} \href{https://github.com/nachiket92/conv-social-pooling}{(link)}, only the model up to lateral and longitudinal manoeuvre prediction has been taken. For training, the model in \cite{deo2018convolutional}, only low and highly congested traffic data is used. The medium congested data is applied for evaluation. For a fair comparison, two models of CS-LSTM \cite{deo2018convolutional} are trained. One CS-LSTM model is trained with human decisions, and the other model with ground truth decisions.

The cases where the human decisions and GT decisions are different have been found. Since ground truth decisions are optimal for both safety and efficiency, these are used for comparison. It has been found that the DST-CAN can predict better decisions than CS-LSTM \cite{deo2018convolutional}. Table \ref{comparison_table} shows a comparison of these results. From the table, it is clear that for the consensus cases, the performance of DST-CAN is slightly better than the CS-LSTM. It has been found that conflicting cases are those where safety is of significant importance. DST-CAN performs much better than CS-LSTM for conflicting cases. For the model trained with GT decisions, there is a $21.15$ $\%$ and 12.12$\%$ improvement in making correct decisions for conflicting cases for lateral and longitudinal cases, respectively. For a model trained with human decisions, there are $5.06 \%$ and $6.37 \%$ improvements for lateral and longitudinal decisions, respectively. Class-wise manoeuvre predictions are also shown to understand the significance of the proposed model. Table \ref{tab:conflict_confusion_mat} shows the different kinds of decisions for conflicting cases for DST-CAN and CS-LSTM models trained with ground truth decisions. Those tables show the number of samples that are classified into the corresponding classes. From the lane-changing manoeuvres, it is clear that DST-CAN performed much better than CS-LSTM. CS-LSTM predicts the decision to follow the same lane for many scenes where it should actually predict changing lanes, as seen from Table \ref{tab:conflict_confusion_mat}. The tables clearly show that DST-CAN predicts better than the CS-LSTM model. Similarly, Table \ref{tab:consensus_confusion_mat} shows decision comparisons for consensus cases. From the tables for consensus cases, it is clear that even with an imbalance in the number of samples for different decision classes, the data-pruning of DST-CAN helps in better performance than CS-LSTM. These results indicate that the proposed DST-CAN accurately predicts safe and efficient lateral and longitudinal decisions.
\section{CONCLUSIONS}\label{CONCLUSIONS}
This paper has presented a novel Deep Spatio Temporal Context-Aware decision Network (DST-CAN) for predictive manoeuvre planning for autonomous vehicles on highways. The surrounding vehicles' past, present, and predicted future trajectories have been encoded into a context and POM grid map. From these encodings, both safe and efficient manoeuvres for an AV are predicted by the predictive manoeuvre decision engine. The decision engine is trained using an imitation learning approach (both human decisions and rule-based ground truth decisions). The performance of DST-CAN is evaluated using well-known US$101$ and I$80$ datasets, and the results are compared  with manoeuvre decisions from CS-LSTM. The performance is compared based on the consensus and conflict scenarios and results clearly indicate that DST-CAN produces a $21.15 \%$ improvement in the number of correct lateral decisions and $12.12 \%$ improvement for correct longitudinal decisions compared to the CS-LSTM approach. Further, an ablation study has been carried out to understand the effect of prediction horizon, effect of human bias and near-collision. DST-CAN provides safe and efficient decisions with minimum near-collisions. One can improve near-collision situations with the inclusion of a global drivable area map in the context-aware grid map. In the future, we plan to incorporate passenger comfort and safety trade-offs along with a global drivable area map for an optimized predictive manoeuvre planning algorithm.

\section*{Acknowledgments}
This work has been funded by the Wipro-IISc Research Innovation
Network (WIRIN), India. 

 
%






\bibliographystyle{IEEEtran}
\bibliography{DST-CAN}

\section{Supplementary material}
The NGSIM dataset is a very highly imbalanced long-tail dataset. A summary of the details of the datasets is given in Table \ref{data_summary}. The table shows that following the same lane and cruising on the same lane is a huge portion of the dataset. If we use the whole dataset for training, the network will overfit these classes. This shows the reason for a data-pruning method for training data generation. The prediction of the poses of vehicles with MNN has uncertainties. We have tested the predictions for different time horizons. Our work defines a grid cell size of $13.846$ feet (approximately $4.22$ meters). Analyzing the table of RMSE for Memory Neuron Network (MNN)-based prediction, we observe a maximum error of $2.74$ meters at a $5$-second prediction horizon. Based on this observation, we assume that the prediction error remains within one grid cell for most time steps. Consequently, we normalize the prediction error at each time step by dividing it by the expected maximum error ($4.25$ meters). This normalized error, represented as $e/4.25$, is an uncertainty measure for the MNN prediction. We then calculate the corresponding certainty level by subtracting the normalized error from $certainty = 1 - (e/4.25)$. Plotting this certainty value with respect to time generates the graph presented in Figure \ref{fig:decayplot} of the supplementary material. However, since this plot only depicts certainty at discrete time intervals (every second), we utilize interpolation to obtain a continuous representation of the certainty function. Equation $6$ in the main text defines the continuous function used to interpolate the discrete certainty values shown in Figure \ref{fig:decayplot}. When working with a new dataset, it is crucial to recalibrate the uncertainty model to reflect the prediction errors associated with that specific data accurately. This involves the following steps:
\begin{enumerate}
    \item Generate predictions for a range of time horizons using a prediction model.
    \item Calculate the Root Mean Squared Error (RMSE) for each time horizon between the model's predictions and the ground truth values in the new dataset.
    \item Plot the RMSE values against time, as shown in the example below.
    \item Interpolate between these discrete points to create a smooth, continuous function representing the uncertainty in predictions over time. This function will be used to model the decreasing certainty in predictions as the prediction horizon extends further into the future.
\end{enumerate}

\section{Comparison with MMnTP for lateral manoeuvre prediction}

The MMnTP approach has recently incorporated the Transformer architecture for multimodal trajectory and lateral manoeuvre prediction, with $k=1,...,6$. Since DST-CAN is a unimodal method $(k=1)$, we compared the results of the DST-CAN model with the unimodal manoeuvre prediction results of MMnTP. Also, DST-CAN predicts both longitudinal and lateral manoeuvres. For comparison with MMnTP, we evaluated only lateral manoeuvres. As shown in Table \ref{Maneuver prediction results}, DST-CAN achieves $93.36 \%$ overall accuracy for lateral manoeuvres for unbalanced unseen testing datasets.  It’s important to note that MMnTP is evaluated on a balanced dataset, but we used an unseen (during training) unbalanced dataset to assess our model. 

For a fair comparison,  we have also made the class-wise maneuver prediction. This decision helps us ascertain if the trained model is skewed towards a particular manoeuvre prediction. We found that DST-CAN yields $69.88 \%$ average accuracy for different manoeuvre classes.  It’s evident that DST-CAN performs well even with unbalanced unseen datasets. The reason for better lateral manoeuvre prediction can be attributed to the better trajectory prediction model of the DST-CAN method for the unimodal case. Since trajectory prediction is better in DST-CAN for unimodal case, it gives better spatio-temporal representation for manoeuvre planning. 

\section{Ground Truth decision generation method}

The ground truth decision generation details are given in Algorithm \ref{GT_generation}.
Here at first, we look if there is any free space on the same lane as the ego vehicle. If there is not enough space, lane change is the next efficient decision. If there is enough space, the vehicle should check whether there is another vehicle in front of it. The next step is to see if that vehicle is slowing down. If it is slowing down, the ego should change lanes to avoid a collision.

\begin{table}[]
\centering
\begin{tabular}{|c|cc|}
\hline
\multirow{2}{*}{\# Models} & \multicolumn{2}{c|}{Models}                    \\ \cline{2-3} 
                           & \multicolumn{1}{c|}{MMnTP}    & DST-CAN (ours) \\ \hline
k=1                        & \multicolumn{1}{c|}{64.60 \%} & 93.36 \%       \\ \hline
k=2                        & \multicolumn{1}{c|}{81.20 \%} & -              \\ \hline
k=3                        & \multicolumn{1}{c|}{86.01 \%} & -              \\ \hline
k=4                        & \multicolumn{1}{c|}{88.04 \%} & -              \\ \hline
k=5                        & \multicolumn{1}{c|}{89.96 \%} & -              \\ \hline
k=6                        & \multicolumn{1}{c|}{90.48 \%} & -              \\ \hline
\end{tabular}
\caption{Percentage of accuracy for manoeuvre prediction}
\label{Maneuver prediction results}
\end{table}

\begin{table}[h]
\centering
\begin{tabular}{lrrr}
                    & \multicolumn{1}{l}{}      & \multicolumn{1}{l}{}      & \multicolumn{1}{l}{}      \\ \hline
                    & \multicolumn{1}{c}{US101} & \multicolumn{1}{c}{US101} & \multicolumn{1}{c}{US101} \\
                    & 07:50-08:05               & 08:05:08:20           & 08:20-08:35          \\ \hline
Follow same lane    & 94.19 \%                  & 95.78 \%              & 95.93 \%              \\
Take left lane      & 3.70 \%                   & 2.67 \%               & 2.91 \%              \\
Take right lane     & 2.11 \%                   & 1.55 \%               & 1.16 \%               \\
Cruise on same lane & 90.49 \%                  & 85.96 \%              & 83.19 \%              \\
Brake               & 9.51 \%                   & 14.04 \%              & 16.81 \%              \\
Total samples       &  1180598                  &  1403095              &  1515240              \\ \hline
                    & \multicolumn{1}{l}{}      & \multicolumn{1}{l}{}      & \multicolumn{1}{l}{}      \\ \hline
                    & \multicolumn{1}{c}{I80}   & \multicolumn{1}{c}{I80}   & \multicolumn{1}{c}{I80}   \\
                    & 16:00-16:15               & 17:00-17:15           & 17:15-17:30           \\ \hline
Follow same lane    & 94.37 \%                  & 95.87 \%              & 96.17 \%              \\
Take left lane      & 4.62 \%                   & 3.60 \%               & 3.19 \%               \\
Take right lane     & 1.01 \%                   & 0.53 \%               & 0.64 \%               \\
Cruise on same lane & 84.58 \%                  & 78.77 \%              & 78.17 \%               \\
Brake               & 15.42 \%                  & 21.23 \%              & 21.83 \%              \\
Total samples       &  1262678                  &  1549918              &  1753791              \\ \hline
\end{tabular}
  \caption{Details of NGSIM traffic dataset}
  \label{data_summary}
\end{table}

\begin{figure}[h]
\begin{center}
\begin{adjustbox}{max size={.9\textwidth}{.2\textheight}}
\begin{tikzpicture}
\begin{axis}
[
xlabel={prediction time horizon (sec)},
ylabel={$ Certainty, \textbf{C}(t)$},
xmin=0, xmax=6,
ymin=0, ymax=1,
]
\addplot+[ycomb, domain=1:5, samples at={1,2,3,4,5}, blue]{0.47+(0.236-(0.04*x))^0.5};
\end{axis}
\end{tikzpicture}
\end{adjustbox}
\caption{Predicted position certainty over different prediction time horizon}
\label{fig:decayplot}
\end{center}
\end{figure}

\begin{algorithm}[h!]
\begin{algorithmic}[1]
    \State \textbf{Inputs}: Social occupancy grid map at current time. $D_S$: number of unoccupied grids on the same lane as ego vehicle at current instant; $D_{pre}$: number of unoccupied grids on the same lane as ego vehicle at 2 seconds ago (for relative velocity of vehicle in front of ego vehicle); $D_{LB}$: \textit{L2} distance to nearest occupied grid in left-backward side; $D_{LF}$: \textit{L2} distance to nearest occupied grid in left-forward side; $D_{RB}$: \textit{L2} distance to nearest occupied grid in right-backward side; $D_{RF}$: \textit{L2} distance to nearest occupied grid in right-forward side; $I_r$: Occupancy of immediate right grid of ego vehicle(1 if occupied, 0 otherwise); $I_l$: Occupancy of immediate left grid of ego vehicle(1 if occupied, 0 otherwise) \;
    \State \textbf{Output}: Ground truth decision at current time \;
    \State $D_{diff}= D_s - D_{pre}$\;
    \If{$D_S > 2$}
        \If{$D_{diff} \ge 0$}
        \State{\quad lateral decision $\leftarrow$ Follow same lane; \newline \phantom a \phantom . \phantom . \phantom .\phantom . \phantom . \phantom . longitudinal decision $\leftarrow$ Cruise;}
        \Else{}
        \If{[$D_{LB}>\sqrt{5}$ \& $D_{LF}>\sqrt{5}$ \& ($D_{RB}<=\sqrt{5}$ \phantom a \phantom . \phantom .       \phantom . \phantom .  \phantom .  \phantom .  \phantom . or $D_{RF}<=\sqrt{5}$ or $I_r=1$) \&       $I_l=0$]}
        \State{\quad lateral decision $\leftarrow$ Take left lane; \newline \phantom a \phantom . \phantom . \phantom .\phantom . \phantom . \phantom a \phantom .\quad longitudinal decision $\leftarrow$ Cruise;}
        \ElsIf{[$D_{RB}>\sqrt{5}$ \& $D_{RF}>\sqrt{5}$ \& ($D_{LB}<=\phantom a \phantom a \phantom a \phantom         a \phantom a \phantom a \sqrt{5}$ or $D_{LF}<=\sqrt{5}$ or $I_l=1$) \& $I_r=0$]}
        \State{\quad lateral decision $\leftarrow$ Take right lane; \newline \phantom a \phantom . \phantom . \phantom .\phantom . \phantom . \phantom a \phantom .\quad longitudinal decision $\leftarrow$ Cruise;}
        \ElsIf{[($D_{RB}>\sqrt{5}$ \& $D_{RF}>\sqrt{5}$) \& ($D_{LB}>\phantom a \phantom a \phantom a \phantom a \phantom a \phantom a \sqrt{5}$ \& $D_{LF}>\sqrt{5}$) \& $I_r=0$ \& $I_l=0$]}
        \State{\quad lateral decision $\leftarrow$ Take right lane; \newline \phantom a \phantom . \phantom . \phantom .\phantom . \phantom . \phantom a \phantom .\quad longitudinal decision $\leftarrow$ Cruise;}
        \Else
        \State{\quad lateral decision $\leftarrow$ Follow same lane; \newline \phantom a \phantom . \phantom . \phantom .\phantom . \phantom . \phantom a \phantom .\quad longitudinal decision $\leftarrow$ Brake;}
    \EndIf
    \EndIf
    \Else
    \If{[$D_{LB}>\sqrt{5}$ \& $D_{LF}>\sqrt{5}$ \& ($D_{RB}<=\sqrt{5}$ \phantom a \phantom . \phantom .          \phantom . \phantom .  \phantom .  \phantom .  \phantom . or $D_{RF}<=\sqrt{5}$ or $I_r=1$) \&           $I_l=0$]}
    \State{\quad lateral decision $\leftarrow$ Take left lane; \newline \phantom a \phantom . \phantom .     \phantom .\phantom . \phantom . \phantom a longitudinal decision $\leftarrow$ Cruise;}
    \ElsIf{[$D_{RB}>\sqrt{5}$ \& $D_{RF}>\sqrt{5}$ \& ($D_{LB}<=\phantom a \phantom a \phantom a \phantom         a \phantom a \phantom a \sqrt{5}$ or $D_{LF}<=\sqrt{5}$ or $I_l=1$) \& $I_r=0$]}
    \State{\quad lateral decision $\leftarrow$ Take right lane; \newline \phantom a \phantom . \phantom .     \phantom .\phantom . \phantom . \phantom a longitudinal decision $\leftarrow$ Cruise;}
    \ElsIf{[($D_{RB}>\sqrt{5}$ \& $D_{RF}>\sqrt{5}$) \& ($D_{LB}>\phantom a \phantom a \phantom a \phantom a \phantom a \phantom a \sqrt{5}$ \& $D_{LF}>\sqrt{5}$) \& $I_r=0$ \& $I_l=0$]}
    \State{\quad lateral decision $\leftarrow$ Take right lane; \newline \phantom a \phantom . \phantom .     \phantom .\phantom . \phantom . \phantom a longitudinal decision $\leftarrow$ Cruise;}
    \Else
    \State{\quad lateral decision $\leftarrow$ Follow same lane; \newline \phantom a \phantom . \phantom . \phantom .\phantom . \phantom . \phantom . longitudinal decision $\leftarrow$ Brake;}
    \EndIf
    \EndIf
\caption{Ground truth generation pseudo-code}
\label{GT_generation}
\end{algorithmic}
\end{algorithm}

\end{document}